%% file: main.tex
\documentclass[dvipsnames]{article}
\pdfoutput=1
\PassOptionsToPackage{numbers, compress}{natbib}

\usepackage[final]{neurips_2021}
\usepackage{latexsym}

\usepackage{natbib}
\usepackage[utf8x]{inputenc} 
\usepackage[T1]{fontenc}    
\usepackage{hyperref}       
\usepackage{url}            
\usepackage{booktabs}       
\usepackage{amsfonts}       
\usepackage{nicefrac}       
\usepackage{microtype}      
\usepackage[table, dvipsnames]{xcolor}  
\usepackage{xcolor}
\usepackage{color}
\usepackage[pdftex]{graphicx}

\usepackage{graphicx}       

\usepackage{caption}
\usepackage{subcaption}

\title{Viola: A Topic Agnostic Generate-and-Rank Dialogue System}

\author{%
  Hyundong Cho, Basel Shbita, Kartik Shenoy, Shuai Liu, Nikhil Patel, \\ \textbf{Hitesh Pindikanti, Jennifer Lee,} and \textbf{Jonathan May}\\
  University of Southern California\\
  Los Angeles, CA, United States \\
  \texttt{\{hjcho, shbita, kshenoy, liushuai, nmpatel,} \\
  \texttt{pindikan, jlee3900, jonmay\}@usc.edu} \\
}

\begin{document}
\maketitle
\begin{abstract}


We present Viola, an open-domain dialogue system for spoken conversation that uses a topic-agnostic dialogue manager based on a simple  generate-and-rank approach. 
Leveraging recent advances of generative dialogue systems powered by large language models, Viola fetches a batch of response candidates from various neural dialogue models trained with different datasets and knowledge-grounding inputs.
Additional responses originating from template-based generators are also considered, depending on the user’s input and detected entities.
The hand-crafted generators build on a dynamic knowledge graph injected with rich content that is crawled from the web and automatically processed on a daily basis.
Viola's response ranker is a fine-tuned polyencoder that chooses the best response given the dialogue history. 
While dedicated annotations for the polyencoder alone can indirectly steer it away from choosing problematic responses, we add rule-based safety nets to detect neural degeneration and a dedicated classifier to filter out offensive content. 
We analyze conversations that Viola took part in for the Alexa Prize Socialbot Grand Challenge 4 and discuss the strengths and weaknesses of our approach. Lastly, we suggest future work with a focus on curating conversation data specifcially for socialbots that will contribute towards a more robust data-driven socialbot. 

\end{abstract}

\input{sections/intro}

\input{sections/nlp_pipeline}

\input{sections/dialogue_manager}

\input{sections/analysis_findings}

\input{sections/discussion_future_work}



\section*{Acknowledgements}
We thank the Alexa Prize organizers for their feedback and advice during the competition.
We would also like to thank our colleagues Kushal Chawla, Wonhyuk Jang, Suji Kim, Yuchen Lin, and Ryan Wei, who provided insights and expertise that greatly assisted our work during the earlier phases of the competition.

\medskip
\bibliographystyle{authordate1}
\typeout{} 
\bibliography{bibtex}

\appendix
\clearpage

\input{sections/appendix.tex}


\end{document}

%% file: sections/intro.tex
\section{Introduction}
\label{sec:intro}

This paper describes \textbf{Viola}, our socialbot for open-domain conversation that was built to compete in the 2020 Alexa Prize Socialbot Grand Challenge. During the competition, users in the United States were able to trigger any one of the socialbots competing in the Alexa Prize at random with a prompt such as \textit{``let's chat''} to converse with it in English and provide a rating and feedback when they decide to end the conversation. 


\input{assets/viola_conversation_example}

Early socialbots such as ELIZA and the winning socialbot of the previous year's competition Emora have shown that well-designed rule-based dialogue systems can be very effective for an engaging conversation  \citep{weizenbaum1966eliza, finch2020emora}. 
While rule-based systems with templated responses have guaranteed high natural language quality and are ideal for traceable and incremental improvements, most of the logic in these systems requires manual effort in order to cover new topics or intents. 
Moreover, the limit of rule-based systems is clear for open-domain dialogue given the innumerable possibilities of how conversations can unravel.  

Hence, we minimize rule-based decisions for Viola and maintain a simple dialogue manager by pushing the limits of end-to-end generative models (henceforth ``generative models'') as the main engine for socialbots in the Alexa Prize competition \citep{khatri2018advancing}. 
Generative models were considered unreliable and still have many known issues of neural degeneration, repetition, inconsistency, and biases, but they have come a long way thanks to large language models and new datasets that try to reduce the aforementioned issues \citep{radford2019language, devlin2018bert, Zhang2020DialoGPTLG, Xu2020RecipesSafety}. 
Also, many recent models and datasets have been shown to be effective for a narrow skillset, such as being empathetic, engaging, or knowledgeable \citep{Rashkin2019TowardsEO, cho2020grounding, Dinan2019WizardOW}.
However, there is only limited success in fluently blending them together \citep{ Roller2020RecipesFB}. 

Recognizing the difficulty for a single model to generate multiple responses that cover a wide variety of potentially good responses for an ongoing conversation, we collect response candidates from multiple dialogue systems and consult a ranker for choosing a suitable response. 
This approach allows us to simplify our system and focus on achieving two major functionalities: 1) make sure that at least one of the response candidates is a good response and 2) make sure that the ranker is able to choose a good response over other inappropriate responses. 
Therefore, most of our contributions are centered around developing a comprehensive suite of response generators and a reliable ranker.  

In practice, Viola is assisted by shallow finite state machines (FSM), additional neural classifiers, and rule-based guardrails to bring out the full potential of generative models and make sure that we abide by the competition guidelines. 
Also, in light of the growing interest in dialogue systems that incorporate structured knowledge from knowledge graphs (KG) \citep{ parthasarathi-pineau-2018-extending, moon2019opendialkg, zhang2020grounded}, we construct a dynamic KG, called the Viola Internal Knowledge Graph (VIKG). 
It is updated daily and follows a semantic model designed to accommodate Viola's needs.   
VIKG retrieves relevant information as input for knowledge-grounded responses and adds templated knowledge responses to the list of response candidates. 



Table \ref{tab:sample_conversation} showcases an example\footnote{In accordance with Alexa Prize rules, we do not share any conversations with actual users. This is a conversation that we conducted with Viola for illustrative purposes.} 
where Viola is able to effectively leverage responses from domain FSMs or various response generators through a poly-encoder ranker \citep{Humeau2019PolyencodersTA}. With the exception of certain intents that can be picked up with high precision and thus be replied by a templated response, all responses from Viola are chosen from a set of responses from neural models and templated responses with information from external knowledge sources. 

Viola attained an average rating of 3.32 for the entire semifinals despite multiple days of low ratings due to technical failures arising from the deployment process of large neural models and GPU-based EC2 instances. 
We believe that our overarching strategy of refining end-to-end models for a two step process of generating response candidates and choosing the best response while relying on minimal handcrafted templates shows a promising path for more dynamic and scalable conversations. 
We identify the pain points applicable to Viola and speech-based socialbots and outline the next steps for improving them.

\subsection{Design Philosophy and Goals}

Based on our overarching strategy and the shortcomings of previous socialbots, we outline desirable design principles and attributes of a socialbot. We built Viola to have a modular and simple pipeline, be mostly data-driven in generating and selecting responses, be responsive to user requests, and maintain a consistent persona. 

\textbf{Modular:} Viola has a modular design that makes it easy to add or update components. With clearly defined inputs and outputs for each component, the modular structure allows us to independently improve components used for feature extraction, response generation, and response selection. 

\textbf{Simple:} We believe that the best process is no process and the best component is no component. While additional components can be added for quick fixes for frequent issues, we ultimately want the underlying system to be reliable enough that extensions are unnecessary. This means inspecting and curating the data that gets fed into our data-driven models and then removing extensions that become redundant. With fewer moving parts, it is easier to identify problems if any and enables faster iterative development.   

\textbf{Responsive:} We develop Viola to be responsive to user requests. While we leverage custom data annotations to take a data-driven approach for a more generalizable and scalable method to be responsive, we also rely on simple rules to make sure Viola is responsive to requests that are specific to Alexa. Users may ask Alexa to play a certain song or invoke another skill, which is not supported in socialbot mode at the moment. 

\textbf{Data-driven:} We minimize hand-crafted rules that can become brittle and difficult to scale and direct our attention to first principles for solving issues by focusing on data. While certain off-the-shelf models are difficult to customize, we examine and modify the training data for components that we know what the training data is and have control over.

\textbf{Persona:} We want Viola to be consistent in its persona such that even though users are aware that they are simply talking to a socialbot rather than a human being, users can feel connected to the socialbot from its unique and interesting persona. We observe that many users are interested in the bot's persona itself and how rich it is, and therefore exceeding their expectations is also aligned with higher ratings. 



\subsection{System Overview}

\begin{figure*}[ht]
    \centering
    \includegraphics[width=\columnwidth]{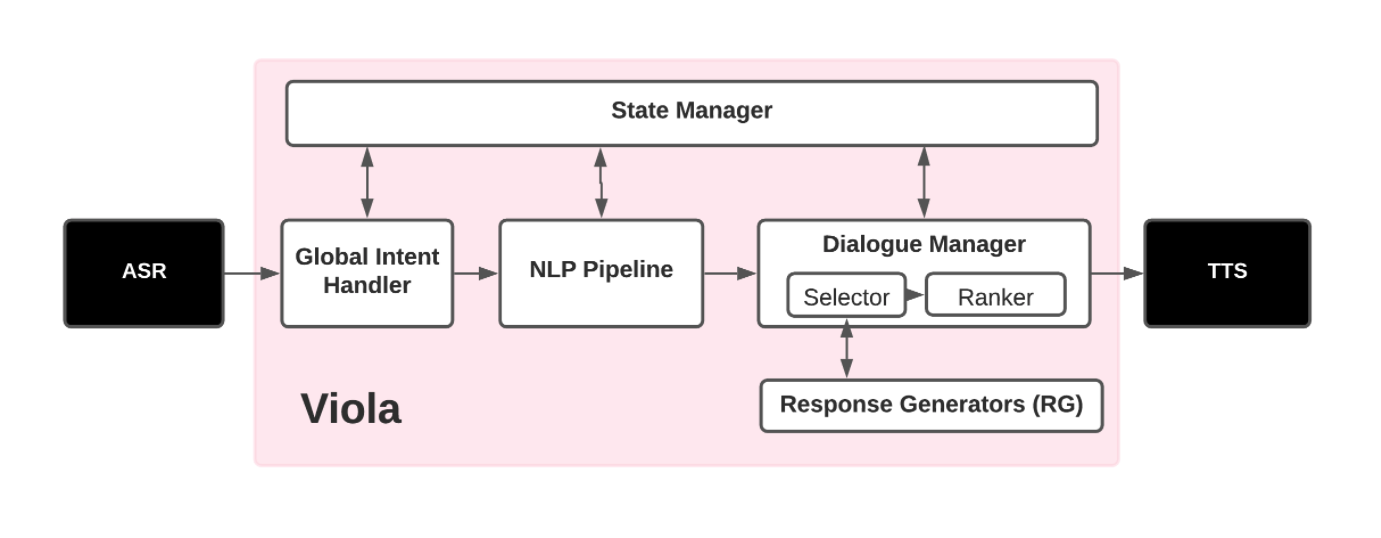}
    \caption{A simplified view of Viola's overall system architecture.}
    \label{fig:sys_arch}
\end{figure*}

Viola is deployed via the Alexa Skills Kit (ASK) as an Alexa skill and it is provided to users as an on-demand service that responds to ASK events. 
ASK events provide the user ID and user's utterance that is transcribed by Amazon's in-house automatic speech recognition (ASR) service. 
The final response generated by Viola is spoken out to the user through Amazon's text-to-speech (TTS) service. 

Viola is built with the Cobot Toolkit Framework and Cortex\footnote{\url{https://cortex.dev}}, a Python package that facilitates the API deployment process for Amazon Web Services (AWS) and Google Cloud Platform \citep{khatri2018advancing}.
Cobot streamlines the socialbot development process through its integration with various services such as AWS Lambda, CodePipeline, DynamoDB, and ECS services. 
It also provides a default socialbot template that has the overall structure shown in Figure \ref{fig:sys_arch} and we build Viola on top of this starting point. 
Deploying internal APIs that is used by Viola is a time-consuming process when going through the CodePipeline that is natively supported by Cobot. 
Therefore, we serve APIs that we update relatively frequently with Cortex. 
With Cortex, we can quickly deploy and update APIs by updating a few configuration files and using its command line interface. 

An overview of Viola's architecture is shown in Figure \ref{fig:sys_arch}. 
The ASR hypothesis of the user utterance is first handled by the global intent handler that makes high level decisions on whether to end the conversation. 
The natural language processing (NLP) pipeline then extracts features from the ASR hypothesis and retrieves information from external knowledge sources that will be used by the dialogue manager. 
The dialogue manager takes the user utterance and information collected from the NLP pipeline to choose which response generators to use, then generates responses, either from neural models or a rule-based template, and selects the best response to be spoken back to the user. We explain each step and components in more detail in the following sections.

%% file: assets/viola_conversation_example.tex
\begin{table}
    \centering
\rowcolors{1}{white}{gray!20}
\renewcommand{\arraystretch}{1.1}
\resizebox{!}{6.2cm}{ 
  \begin{tabular}{ | m{3.2cm} | p{10cm} | }
    \hline
    \textbf{Source} & \textbf{Utterance} \\ \hline
    USER & let’s chat \\ \hline
    LAUNCH & Hi, this is an Alexa Prize Socialbot. simpson, It's good to have you back! How have things been for you since we last talked? Did you do anything interesting or have things been pretty much the same? \\ \hline
    
    USER & its pretty much the same but i am going to las vegas this weekend \\ \hline
    BLENDERBOT & That sounds like a lot of fun! I've never been to Vegas, but I've always wanted to go. Have you been there before? \\ \hline
    
    USER & no this will be my first time do you know some popular places in vegas \\ \hline
    SPOLIN & Well I know a couple places. I know the casinos and the strip. But I don't know much about Vegas.\\ \hline

    USER & cool no problem do you know any movies \\ \hline
    MOVIES\_RESPONSE & ive been watching a lot of movies lately. have you seen hamilton? \\ \hline
    
    USER & no have you watched deadpool two \\ \hline
    MOVIES\_RESPONSE & ha! I love that movie. did you know that the comic books depict domino as a long-time ally and former girlfriend of cable, who hates deadpool. in the film, she is introduced as an ally of deadpool working against cable. which character did you like best? \\ \hline

    USER & lets talk about sports instead    \\ \hline
    DIALOGPT\_ALEXA & Ok. I like sports. Do you like sports? \\ \hline
    
    USER & yes do you play football \\ \hline
    VIKG-RG & that reminds me, my robot cousin informed me that NFL star Alvin Kamara has a 75 million contract, but hasn't spent a dime of his football earnings \\ \hline
    
    USER & oh okay who is better messi or ronaldo \\ \hline
    SPOLIN & Ronaldo, he's not very good, but he has a lot of heart. \\ \hline
    
    USER & hmm i think messi is great at team play and ronaldo is great in dribbling skills\\ \hline
    NRG & That's cool. Neymar has the potential to be the next super star as long as he does less diving in the game. Mbappe also has huge potential. He is incredibly fast and has a lot of talent \\ \hline
    
    USER & yes thats true they both are very good do you know what is neymars age \\ \hline
    QA & Neymar is 29 years old. He was born on February 5, 1992. \\ \hline

    USER & goodbye \\ \hline
    RULE-BASED & I think you want to end the conversation. Just say stop so I can exit socialbot mode. \\ \hline
    
    USER & stop \\ 

    \hline
  \end{tabular}

  }
    \caption{An example conversation from Viola that shows how it flexibly chooses response from various sources. Viola's response selector chooses the most appropriate response among those generated from neural models and templated responses.}
    \label{tab:sample_conversation}
\end{table}

%% file: sections/nlp_pipeline.tex
\section{NLP Pipeline}
\label{sec:nlp}

Viola's NLP pipeline is executed for every utterance that makes it through the global intent handler. 
It has a flexible design that allows modules to run in parallel and also in sequence based on any dependency among them. 
Since our overarching strategy is to minimize intermediate rule-based decisions and use generative models, we keep our NLP pipeline simple compared to dialogue systems that mainly rely on robust intent classification and dialogue act classification to fill in hand-crafted templates. 
Our NLP pipeline consists of modules that are necessary for retrieving knowledge that grounds generative models and extracting features for rule-based guardrails that remove problematic response candidates. 

\subsection{Named Entity Recognition (NER)}
\label{sec:nlp_ner}

Identifying entities is important for socialbots since the conversation often centers around them.
Entities include both named entities (i.e., persons, locations, organizations) and key noun phrases (e.g., cars).
The entity recognition component is the default module provided to us as part of Cobot's NLP pipeline.
It is based on a trained BERT-based model for named entity recognition on the TopicalChat Dataset and Alexa Prize data from previous years \citep{devlin2018bert, gopalakrishnan2019topical}.
The model tags entities with entity types from domains including sports, books, music, fashion, etc.

\subsection{Punctuation}

Alexa's ASR service does not provide punctuation or segmentation information of the user's original speech input.
Punctuation is important for downstream generative response generators because they were trained with text data that had punctuation and they are not robust to input without it. 
Cobot provides a default punctuation model as an API.

\subsection{Intent classification}

While the default intent models provided by the Alexa Prize are good starting points for intent-based dialogue management, they cannot be customized to add new intents. 
Since we mostly take an end-to-end approach with response generation, intent classification is not as important as it would be for hand-scripted conversation flows that require accurate intent classification to guide the dialogue manager. 
Instead, we use intent classification for specific situations where neural response generators fail or scenarios that occur as part of the unique nature of the Alexa Prize competition that the data-driven models have not seen as part of their training. 
Given our objective with intent classification, we want a light-weight and easily extensible intent classifier that can be quickly retrained and deployed.

We use Snips NLU as our main engine for classifying intents \citep{coucke2018snips}. 
Snips NLU uses simple regular expression matching as a first stop to match against intents from training and logistic regression if there is no direct match. For logistic regression, TF-IDF and co-occurence features  are extracted from utterances and a classifier is trained with simple stochastic gradient descent.
It is very easy to add new training examples and custom intents by configuring YAML files, but the model  can be brittle---classification results can change based on the presence of seemingly irrelevant features such as the presence of certain punctuation.



\subsection{Information Retrieval from the Viola Internal Knowledge Graph (VIKG)}
\label{sec:nlp_vikg}

We have constructed a knowledge graph, called the Viola Internal Knowledge Graph (VIKG), with AWS Neptune as our graph database service.
VIKG serves as a knowledge retrieval module to power several template-based response generators and domain-specific conversations.
The knowledge graph stores millions of different topics, entities, news stories, public opinions, and additional domain-specific (e.g., movies) data and factual knowledge.

\begin{figure*}[ht]
    \centering
    \includegraphics[width=\columnwidth]{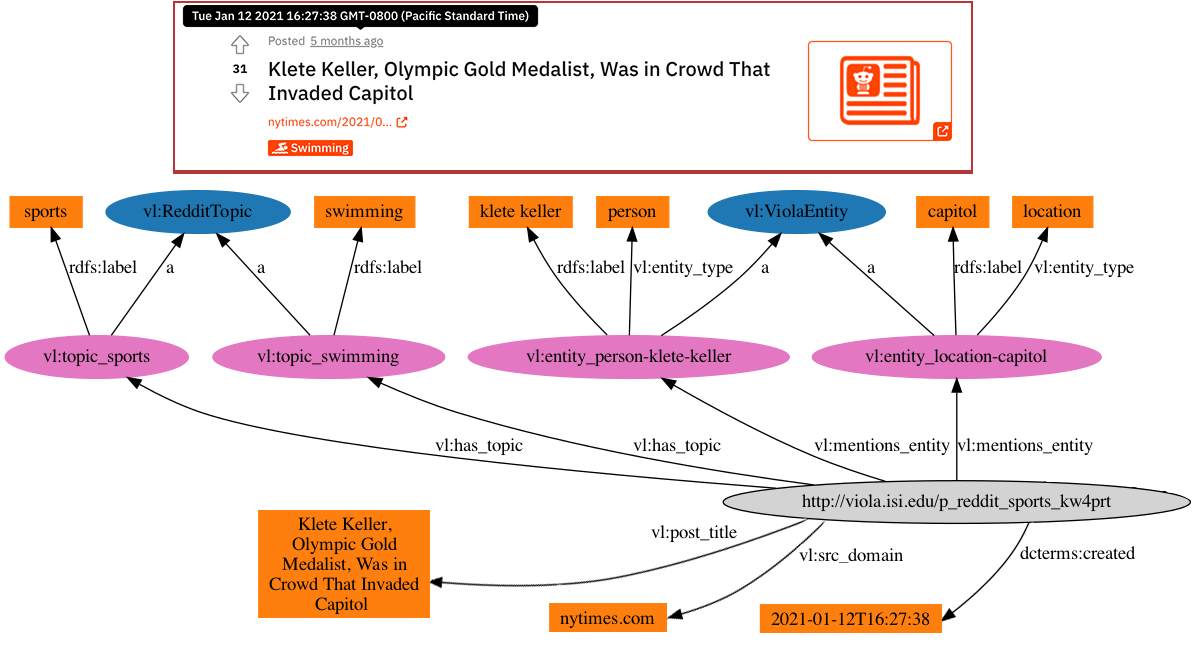}
    \caption{A snapshot of a Reddit post that was crawled (top), processed and then stored in VIKG (bottom). In the graph database, the post instance identifier node is seen in gray, topic and entity instances in pink, classes in blue, and literal values in orange.}
    \label{fig:vikg_example}
\end{figure*}

VIKG is constructed to describe applicable classes and properties, represented as nodes and edges, following a predefined semantic model we designed.
We leverage open data sources such as Reddit, IMDB, and the Washington Post to populate our knowledge graph.
The data is crawled on a daily basis, then filtered and cleansed automatically to avoid profanity and inadequate content.
We then process the data using linguistic and statistical methods we developed with \textit{SpaCy} \citep{honnibal2017spacy} and annotate it with the entity and topic metadata using Cobot’s entity tagger (section \ref{sec:nlp_ner}) and topic classifier, which is based on a model trained on the same data as the entity tagger.
In our implementation, the RDFLib\footnote{\url{https://rdflib.readthedocs.io/}} is used to construct the graph triples and convert the data to a semantic graph according to the resource description framework (RDF).
Figure \ref{fig:vikg_example} shows one example of a Reddit post and how it is stored in VIKG with its corresponding annotations.

Once we receive the user’s utterance and additional metadata originating from other pipeline components (i.e., entities) we leverage the SPARQL query language\footnote{\url{https://www.w3.org/TR/sparql11-query/}} to query relevant instance data and identifiers from VIKG.
The retrieved data is used by domain-specific and template-based response generators (sections \ref{sec:domains_fsms} and \ref{sec:rg_vikg}).

\subsection{DBpedia information retrieval}
\label{sec:nlp_dbpedia}
Viola's knowledge-grounded response generators require external knowledge as input to their models (see section \ref{sec:rg_kn}).
To power these generators, we fetch relevant information from DBpedia \citep{auer2007dbpedia} about the entities suggested by the NER pipeline component mentioned in section \ref{sec:nlp_ner}, thus generating sentences related to the dialogue context.

%% file: sections/dialogue_manager.tex
\section{Dialogue Manager}
\label{sec:dialogue_manager}

\begin{figure*}[ht]
    \centering
    \includegraphics[width=\columnwidth]{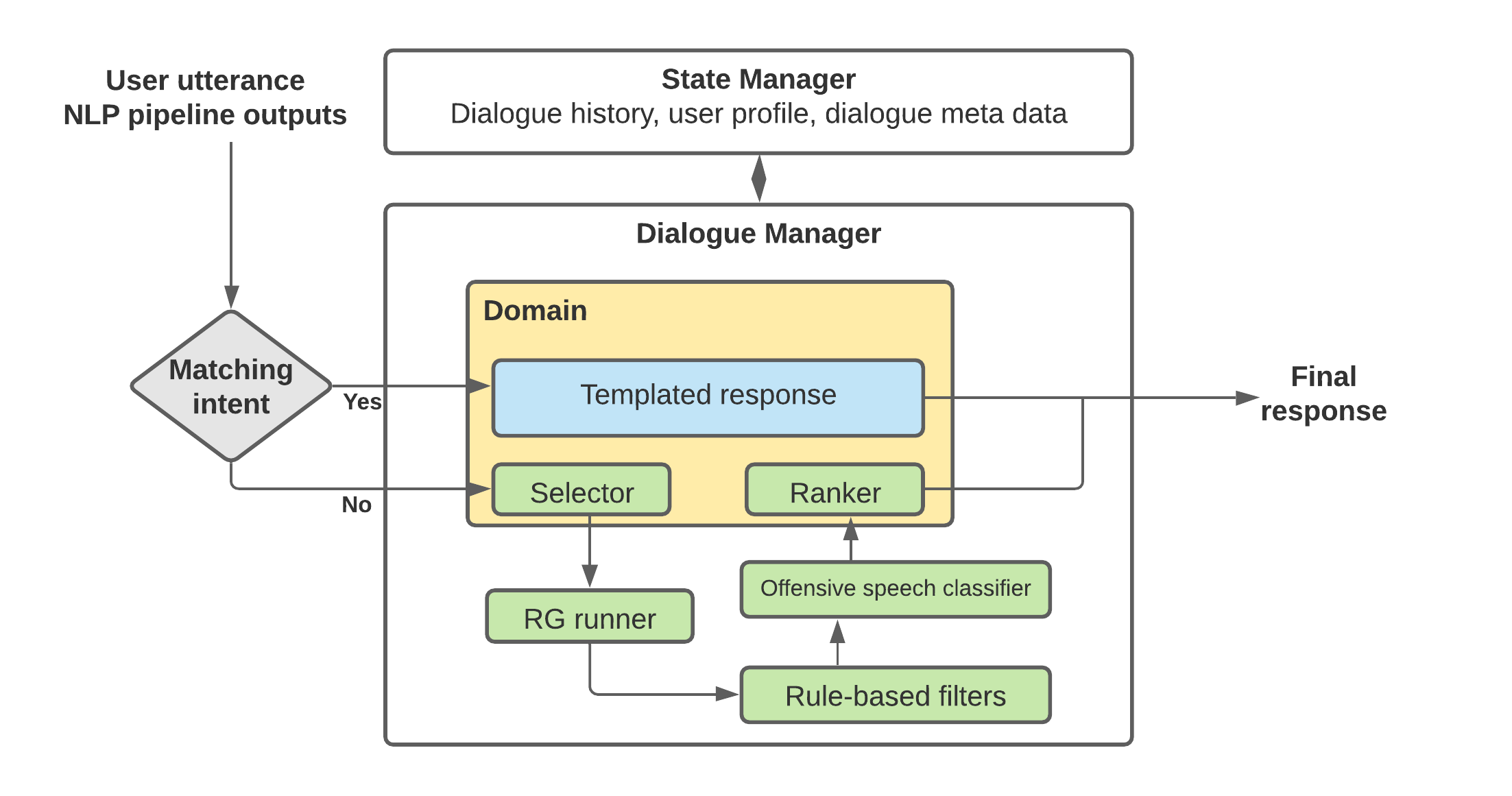}
    \caption{Viola's dialogue manager overview.}
    \label{fig:dm_overview}
\end{figure*}

An overview of Viola's dialogue manager (DM) is shown in Figure \ref{fig:dm_overview}. 
The DM handles the high-level logic of the conversation flow and is the glue that integrates Viola's components together.
Unless we have high conviction from intent classification to use a templated response from a domain specific FSM, the DM's default strategy is to generate responses from all available response generators and then let the ranker choose the best response that will get passed on to the TTS service.

The DM leverages the relevant context from the output of the NLP pipeline (section \ref{sec:nlp}).
At each turn, it is responsible for handling the logic, storing metadata, generating responses (section \ref{sec:response_generators}),
employing guardrails to eliminate inadequate content and repetition (section \ref{sec:domains_guardrails}),
executing domain-specific flows if applicable (section \ref{sec:domains_fsms}), and choosing the appropriate response  (section \ref{sec:response_ranker}).

\input{sections/response_generators}

\subsection{Guardrails}
\label{sec:domains_guardrails}

While we amend the training data to make sure that the neural response generators do not generate problematic responses, some of these responses can be more easily and directly handled by simple rules and off-the-shelf systems. 
Viola employs a number of low-latency guardrails whose execution handles multiple functions such as detecting profanity, repetition, and other specific issues of neural response generators. We use the following guardrails to ensure smooth functioning of the bot:

\textbf{Avoiding repetitive responses}: We maintain a history of the bot responses by storing the embeddings of all the bot responses as well as the embeddings of the last sentences of all the bot responses. We generate these embeddings using Sentence-BERT \citep{reimers-2019-sentence-bert}. When we have all the bot responses generated by the response generators, we compare both the embeddings of the entire text and the last sentence of all the bot responses with these stored embeddings to avoid repetition of bot responses, thereby ensuring diversity in the responses that might be selected.

\textbf{Handling offensive and sensitive utterances}: We found that the default offensive speech classifier provided with Cobot was too sensitive and often incorrect, classifying utterances such as ``\textit{i kicked his butt in Mario Kart}" and ``\textit{justin you}" as offensive. 
We replace the default classifier with an off-the-shelf library Detoxify \citep{Detoxify} to determine the offensiveness of the utterances as we saw that it attains a better accuracy on our internal test set. 
The library uses a model that has been fine-tuned from the Uncased BERT-Base \citep{DBLP:journals/corr/abs-1810-04805} model using the dataset from Kaggle's 2018 Toxic Comment Classification Challenge \footnote{\url{https://www.kaggle.com/c/jigsaw-toxic-comment-classification-challenge/data}}. First, we use this profanity handler to determine the offensiveness and toxicity of the user utterance. If the user utterance crosses a predetermined threshold, we classify it as offensive and generate a templated bot response indicating the bot's discomfort on hearing this. We also use this profanity handler to check potential bot responses for profanity and, using the same threshold, eliminate offensive responses.

\textbf{Detecting neural degeneration}: Neural models often hallucinate and generate repetitive $n$-grams. In order to detect these cases and remove them from responses to be ranked, we count the occurrences of $n$-grams in each of the responses generated. We then compare these counts with predetermined thresholds and filter out the responses that cross these limits.

\textbf{Filtering responses not suitable for socialbots}: For each response that is generated, we also determine its intent using the pretrained Snips NLU model.
We have a specific intent that covers cases where the responses describe actions and assets like a human. For example, we do not expect a socialbot to say ``I just got back from getting groceries for dinner" as it will not eat nor cook. 
We expand our Snips NLU training files to keep track of these types of utterances. 
If our responses have this intent, we filter them out. We hope that the polyencoder takes care of this but we add it as an additional safety measure for frequently occurring cases.

\subsection{Domains as Finite State Machines}
\label{sec:domains_fsms}

In Viola, there are several predefined domains (i.e., topics of interest) corresponding to mini dialogue managers that override the behavior of the default DM and can prefer specific response candidates over others or override them.
These domains are \textbf{Movies}, \textbf{Sports}, \textbf{Music}, and \textbf{News}.

Each predefined domain corresponds to a dialogue flow that can be viewed as a sequence of user/bot activity states and transitions between them.
Thus, the dialogue control for each domain is formulated as a FSM that manages a short dialogue segment.

\begin{figure*}[ht]
    \centering
    \includegraphics[width=0.75\textwidth]{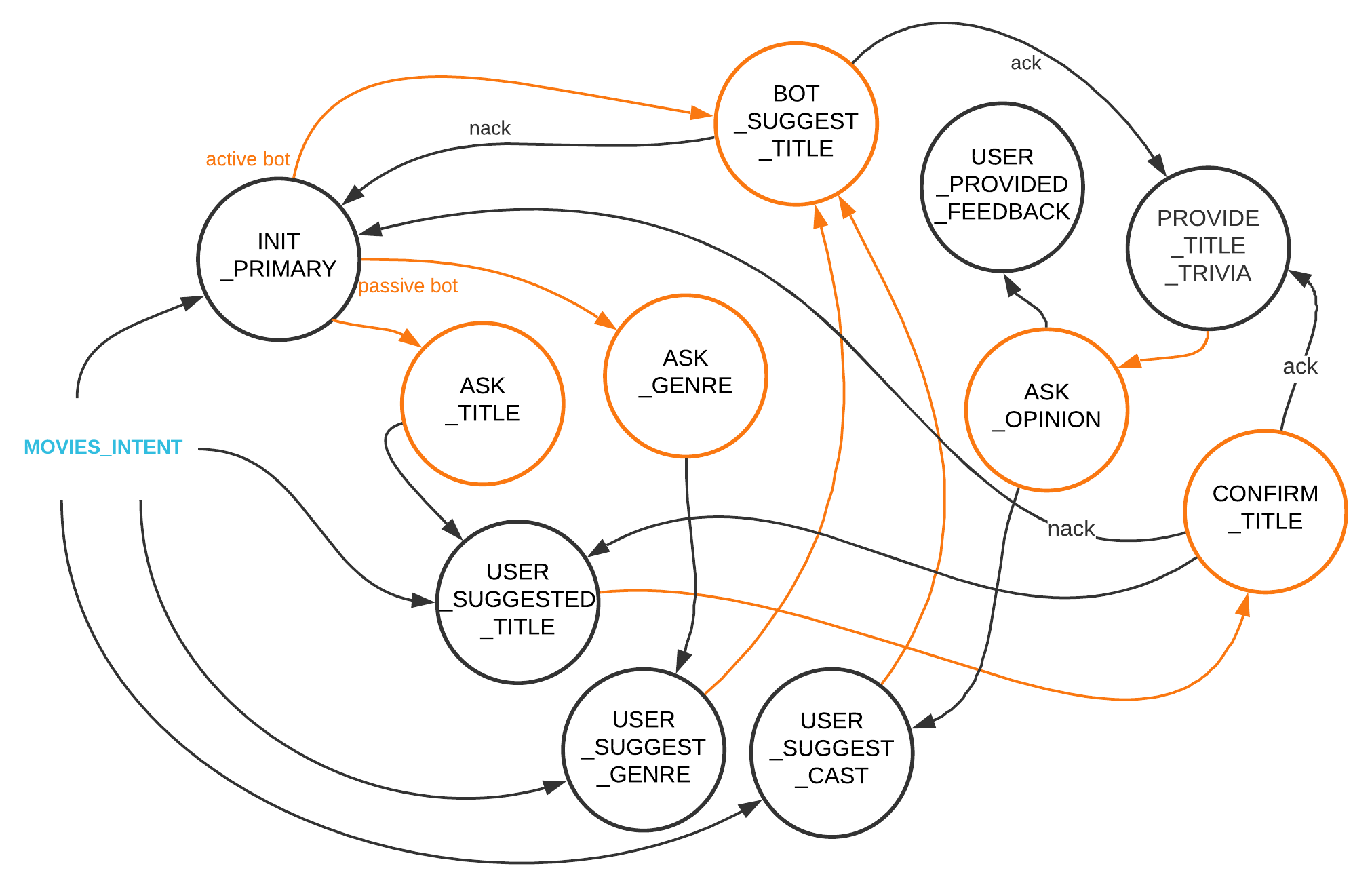}
    \caption{A snippet of the Movies domain FSM. The \texttt{MOVIES\_INTENT} shown on the left in blue is the entry point to the domain. Ingress (user) states are shown in black. Egress (bot) states are shown in orange. Black edges represent user input. Orange edges represent bot response.}
    \label{fig:fsm_movies}
\end{figure*}

\begin{figure*}[ht]
    \centering
    \includegraphics[width=0.6\textwidth]{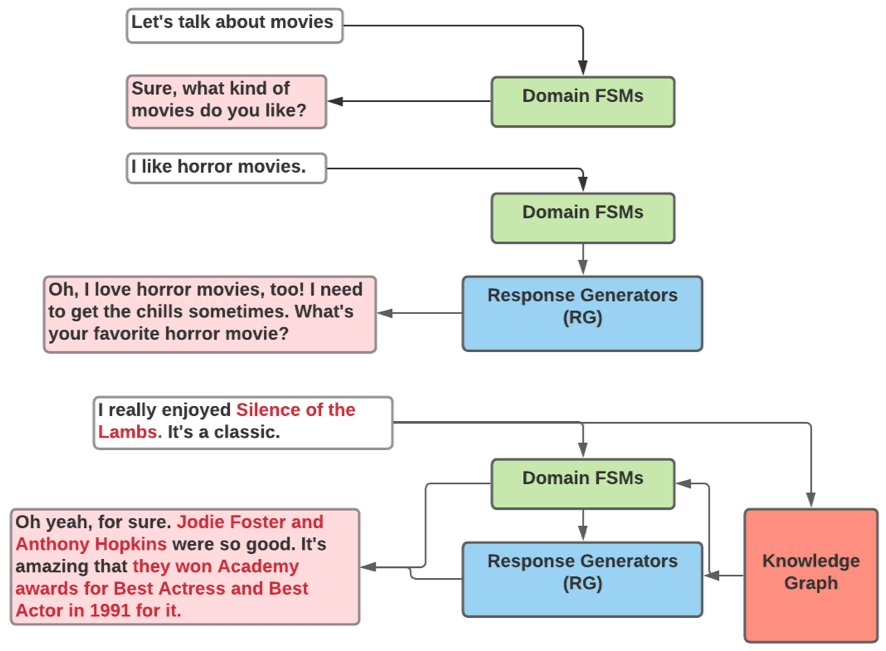}
    \caption{An excerpt from a sample subdialogue between the user and Viola, powered by the Movies domain FSM and the Viola Internal Knowledge Graph.}
    \label{fig:chat_movies}
\end{figure*}

The inventory of states is organized based on the primary dialogue intents (determined by features originating from the NLP pipeline) of the bot at a turn in the envisioned dialogue flow of the domain.
In each turn, two states are dictated: ingress and egress.
The ingress state is determined by the user’s input in the current turn and by metadata from the previous turns.
The egress state is a response state to the ingress state; it is randomly picked by the bot in some cases.
As an example, Figure \ref{fig:fsm_movies} shows the FSM of the \textbf{Movies} domain.
The nodes shown in black are ingress states, and the ones in orange are egress states.
Each user input is mapped to an incoming edge into an ingress state (black edges).
Each transition from an ingress to an egress state (orange edges) generates a response from the bot and concludes a single turn.
For example, in Figure \ref{fig:fsm_movies}, the state \texttt{INIT\_PRIMARY} means that the user has shown general interest in talking about movies (with no explicit mention of a title, cast or genre).
The bot has several options (three outgoing orange edges).
It randomly picks one.
It could take an active approach and suggest a movie to the user (and thus transition to the state \texttt{BOT\_SUGGEST\_TITLE}), or it could take a passive decision and instead decide to ask the user about his favorite genre (and thus transition to the state \texttt{ASK\_GENRE}, as demonstrated in the first response utterance in the conversation example in Figure \ref{fig:chat_movies}).

The FSM's response generator is based on predefined templates.
Each egress transition generates a relevant response and stores metadata on the state manager for future use by the FSM.
It further stores state history, possible topics, and entities to discuss, including whether a topic has been discussed, suggested by the bot, or mentioned by the user.
Each template is crafted carefully to construct the bot's response in a granular way.
The response is a template that eventually gets populated with content.
Content can be a topic, some information from a knowledge source (e.g., VIKG), part of the previous user utterance to be echoed, etc.
Figure \ref{fig:chat_movies} shows a simple conversation consisting of three turns between the user and the bot.
As seen in the last utterance in Figure \ref{fig:chat_movies}, the user's input, which includes a movie entity (\textit{Silence of the Lambs}), invokes a proper template that gets further populated and extended with data fetched from our knowledge graph.
In this case, it is about the cast and an interesting fact about them.

While the bot has a default dialogue flow for the domain, the DM still allows the user to take the initiative at any time and ``steer away'' from the FSM.
Suppose the user's input does not correspond to any transition for the current state.
In that case, the default behavior of the DM is triggered.
This allows either initiating a new FSM or falling back into the general generate-and-rank mechanism with a subset of candidate response generators applicable in this case.
The specific subset is determined by the current state and the custom domain DM.

Each one of the predefined DMs implements its own logic and FSM.
The Music, Sports, and News DMs have similar logic and infrastructure to the Movies domain.
There are 13 states in the Movies domain, 18 states in Music, 46 in Sports, and 18 in the News domain.
For each state, there is at least one available transition.
At least one template for each egress transition can be used, enabling a vast amount of possible variations in the bot's responses and a more significant number of different conversation flows in these domains.

\input{sections/response_ranker}

%% file: sections/response_generators.tex
\subsection{Response Generators}
\label{sec:response_generators}

Our suite of response generators (RGs) mainly consists of (1) end-to-end RGs, which only need the dialogue history and the current incoming utterance as text, (2) knowledge-grounded RGs, which take a two-step process of retrieving relevant knowledge based on the user utterance and using it to generate a response, and (3) rule-based RGs, which use templated responses for frequently occurring situations that we can determine with high precision. We describe each RG in detail in this section and discuss the results in \autoref{sec:analysis}. 

\subsubsection{End-to-End RGs}

\textbf{Topical neural RG (NRG)} is a dialogue model that is provided by the Alexa team as an API as part of Cobot. 
It is a Transformer-based model that is trained with the Topical-chat dataset and also conversations that took place within the Alexa Prize \citep{gopalakrishnan2019topical}. 
The main advantage of this RG is it is a model that learned from conversations that were built with carefully designed conversation flows by previous competitors and is generative, maintaining some degree of generalizability to unseen utterances. 
While the exact training data for the underlying model is not known to us, it seems to be quite robust to noisy transcribed utterances as it is trained with previous Alexa conversations. 
It also rarely generates responses that demonstrate neural degeneration, possibly learning from hand-crafted templated responses that are very natural. 
It uses stochastic decoding and therefore every call to the API with the same input returns a different result. 
We make $N=5$ parallel calls to get a diverse set of results. We set a time limit of one second and use only the responses that we get back within this limit. 

\textbf{Policy-driven Topical NRG (PD-NRG)} is another  Transformer-based dialogue model \citep{Hedayatnia2020PolicyDrivenNR} that is provided by Cobot. 
The main advantage of this RG is that it uses a dialogue policy to plan the content and style of the target responses in the form of an action plan. 
It uses the dialogue act classification result as an intermediate input to ground its responses. 
This allows the creation of informative and engaging responses. 
We use this service the same way as the NRG model in making parallel calls and setting a time limit. 

\textbf{SpolinBot} is a DialoGPT model, fine-tuned  with the SPOLIN dataset \citep{Zhang2020DialoGPTLG, cho2020grounding}. 
DialoGPT itself is a GPT-2 model \citep{radford2019language} that is fine-tuned with Reddit threads.  
Fine-tuning is done with autoregressive language modeling loss where the turns that are to be modeled are given a weighting of 1.0 while the turns that are ignored, usually the user utterances, are given a zero weighting. 
The fine-tuning process is identical for all DialoGPT-based response generators.

SPOLIN consists of about 68,000 utterance pairs that are called \textit{yes-ands}. \textit{Yes-ands} are utterance pairs in which the response abides by the ``\textit{Yes-and}'' principle: a rule of thumb of improvisational theatre (improv) that encourages actors to agree with each other about the current scene and carry it forward with relevant new information in order to quickly construct a common world view. By being fine-tuned with \textit{yes-ands}, SpolinBot is trained to ground its response to the previous turn and generate engaging \textit{yes-and} responses that aim at building common ground with the user.

The initial set of \textit{yes-ands} in SPOLIN were transcribed by annotators from a long-form improv podcast called \textit{Spontaneanation}\footnote{\url{https://www.earwolf.com/show/spontaneanation-with-paul-f-tompkins/}} where actors improvise with a scene given by the listeners as the prompt. The dataset was expanded by training a bootstrapped BERT classifier with the initial set to find potential \textit{yes-ands} from existing dialogue corpora, specifically the Cornell Movie-Dialogs Corpus and SubTle \citep{danescu2011chameleons, ameixa2013subtitles}. Candidate \textit{yes-ands} were validated by annotators before being added to SPOLIN.

\textbf{DialoGPT + Alexa:} We also use an indirect reinforcement learning approach with a DialoGPT model that is fine-tuned with Viola's conversations that received a rating of 4.0 or above or lasted more than 20 turns. 
We saw that there were many conversations that were given a high rating despite lasting fewer than five turns so we leave them out of training.
We also filter out utterances coming from rule-based response generators. 

\textbf{DialoGPT + GPT-3 alternative responses (DialoGPT-3): } 
DialoGPT + Alexa is a RG that is fine-tuned with conversations with high ratings, while DialGPT + GPT-3 takes advantage of the conversations that received low ratings. 
We consult OpenAI's GPT-3 API, which serves a multi-billion parameter language model, for alternative responses that could potentially be better options \citep{brown2020language}. 
For conversations that received a 1.0 rating, we generated up to 150 words as the alternative responses with GPT-3's \textit{davinci} model for each of the bot's turn using up to 100 words from the dialogue history as the context. 
We cap our training data to 20K lines for this model and use it to fine-tune another DialoGPT model. Some cherry-picked and lemon-picked examples of alternative responses provided by GPT-3 are shown in the appendix (Table \ref{tab:gpt3-good} and \ref{tab:gpt3-bad}). 



\textbf{BlenderBot} is a sequence-to-sequence Transformer model that was trained to blend multiple skills together \citep{Roller2020RecipesFB}. 
It is trained with the Blended Skill Talk dataset, which is a multi-task conversational dataset created by combining the Empathetic Dialogues, Wizard of Wikipedia, and Persona-chat dataset \citep{smith2020can, Rashkin2019TowardsEO, Zhang2018PersonalizingDA, Dinan2019WizardOW}. 
Due to computational constraints, we use the 360 million parameter model that is a distilled version of the 2.7 billion parameter model.

\subsubsection{Knowledge-grounded RGs}
\label{sec:rg_kn}

\textbf{Question-Answering RG (QA-RG):} To answer factual questions asked by users, we take advantage of the EVI API that is provided by Cobot. 
It is a QA service that returns responses that you would get from asking questions to Alexa devices.
EVI tends to interpret many non-questions as questions and return strange responses, so we use a question classification service that is also provided by Cobot to make sure that the incoming utterance to the QA-RG is in fact a question.
If it is considered a question by the question classifier, QA-RG uses EVI's response. 

\textbf{NRG + knowledge \& PD-NRG + knowledge:} NRG and PD-NRG are able to take an additional \textit{knowledge} parameter as input to ground its response. 
These model variants are only used when there is some knowledge retrieved based on named entities in the user utterance.
Knowledge should be provided as natural text and it is appended to the dialogue history \citep{gopalakrishnan2019topical}. 
The underlying Transformer model learns to attend to the dialogue history and knowledge span to generate a response. 
However, it is not guaranteed that the given knowledge will be incorporated into the final response, which is in line with the experimental results for the Athena dialogue system in the previous year's Alexa Prize  \citep{harrison2020athena}.  

\textbf{VIKG (template-based) response generator:} \label{sec:rg_vikg} To enable the delivery of factual and opinionated up-to-date world knowledge in a conversational style, we implement a template-based response generator that builds on Reddit headers (originating from VIKG data we acquire in the NLP pipeline as described in section \ref{sec:nlp_vikg}).
The provided \textit{knowledge} relates to topics and entities detected in the current user utterance.
The responses generated by this module are ranked using a score that consists of several parameters: named entities (higher overlap of entities with the current user utterance is preferred), date (recent data is preferred) and topic (topic match with current user utterance is preferred).
For example, the user utterance \textit{``i just got back from a trip in \underline{japan} and \underline{china}} (detected entities are underlined) results in the response \textit{``woah, related to that, my bot friend told me that \underline{Japan urges Europe to have stronger military presence in Asia to counter China}''} (data originating from VIKG is underlined.)
\subsubsection{Rule-based RGs}

\textbf{Launch RG} For the first few turns of the conversation, the Launch RG takes full control and uses templated responses to initiate the conversation. 
If we have the name of the user from a previous conversation, the Launch RG will use it to greet the user, saying that we were looking forward to having him/her back and asking about any interesting personal events that happened since the last conversation. 
If the user is a first timer, we will ask for their name and store it for that user profile. Names are detected based on Snips intent results and regular expression matching. 
While many previous socialbots tend to direct the conversation towards topics that they are comfortable with, through the Launch RG, Viola gives the user a chance to take more control over what to discuss and only moves on to use models such as NRG to suggest new topics if it seems the current topic has reached a dead end.  

\textbf{Favorite RG: } In order to maintain a consistent persona, we use basic regular expression matching to pick up on utterances that asks Viola about its preferences. For each matching regular expression, we have a templated response that answers what Viola prefers and also why. 

\textbf{Fallback RG: }
When the ranker chooses an inappropriate response and the user gets confused, users convey this confusion with multiple \textit{``what''} utterances or say things such as \textit{``alexa I think you’re crazy''}. 
They may also say \textit{``you just said that''} when the bot is being repetitive or \textit{``you just said you liked to go hiking''} when the bot makes a contradictory statement.
While we want the generative models to be able to understand these utterances and redirect conversations in a sensible way, we often see them go on a tangent and use Viola’s own previous utterances to continue a conversation rather than listening to the user. 
Therefore, we set up custom Snips NLU intents and regular expressions to capture these cases and map them against multiple templated responses that get chosen at random to make sure that we are responsive to these instances. 

\textbf{Sensitive RG: } 
In accordance with the Alexa Prize rules, there are sensitive questions and utterances from the user that socialbots should either defer or answer with a templated response in order to pass for certification as a valid Alexa skill. 
Any questions that ask for medical, financial, or legal advice should not be answered and a direction to consult an alternative source, such as human experts, for such advice should be provided. 
Distress signals and questions about user privacy are answered with standard responses provided by the Alexa Prize team. 
Like the Fallback RG, the sensitive RG has Snips NLU intents mapped to templated responses.

%% file: sections/response_ranker.tex
\subsection{Response Ranker}
\label{sec:response_ranker}

Since Viola generates multiple response candidates, a ranker is needed to choose the best one as the final response to be passed on to the TTS service.
Here we list the different type of response rankers that we used throughout the Alexa Prize competition.

\subsubsection{Conversation Evaluator}

The Conversation Evaluator is provided by Cobot as an API for evaluating a response given the two most recent turns from the conversation. 
It takes in a batch of conversation context and response pairs and estimates the probability of the responses falling into each of the five classes: \texttt{is\_response\_on\_topic}, \texttt{is\_response\_comprehensible}, \texttt{is\_response\_interesting}, \texttt{response\_engages\_user}, and \texttt{is\_response\_erroneous}. 
The probabilities are on the scale of 0 to 1. 
We use the \texttt{is\_response\_erroneous} metric as it is the one that was recommended as the metric that best aligns with overall human judgement for whether a response is good or bad.
For the \texttt{is\_response\_erroneous} metric, lower scores indicate that the response is more likely to be appropriate for the given dialogue history.

\subsubsection{Poly-encoder}

The poly-encoder is a state-of-the-art model for multi-sentence scoring, which scores label sequences with a given context sequence \citep{Humeau2019PolyencodersTA}. 
In our case, the label sequences are candidate responses, and the context is the conversation history. The poly-encoder encodes the context and the candidate with two separate transformers. 
The candidate is encoded into a single vector, while the context is encoded into $m$ vectors since the context is usually much longer than the candidate. 
Then, the candidate vector is used as the query to attend over the context vectors to get a single vector representation of the context, and the final score of the candidate is the dot product of the candidate vector and the single context vector.

The default poly-encoder model is pre-trained on the Reddit dataset using standard masked language modeling and next-utterance objectives, and is then fine-tuned on the ConvAI2 dataset for the multi-sentence scoring task \citep{mazare-etal-2018-training, 10.1007/978-3-030-29135-8_7}.
This version of poly-encoder is reported to have an 89+\% accuracy (hits@1/20) on ConvAI2, but similar performance should only be expected on data close to the ConvAI2 dataset. 
ConvAI2 is a conversation dataset based on Persona Chat, in which each dialogue comes with personas of the two speakers which are described by short sentences, and the speakers are asked to get to know each other through the conversation \citep{Zhang2018PersonalizingDA}.

The setting of the ConvAI2 dataset is similar to ours in terms of engagement, but there are still some differences between them. 
The dialogues in the ConvAI2 dataset are between humans and conditioned on the persona of the speakers, but in our case, the conversation is between a human and a bot and has no persona information. 
Also, the domain of the ConvAI2 dataset is relatively narrow compared with our social bot setting. 
Therefore, we create a new dataset for poly-encoder fine-tuning by manually annotating some selected conversations between users and our bot and fine-tuning the poly-encoder model with the new data. 
More detail of the annotation and fine-tuning process are discussed in the following sections.

\subsubsection{Fine-tuning the Poly-encoder}
\begin{figure*}[h!]
    \centering
    \includegraphics[width=\textwidth]{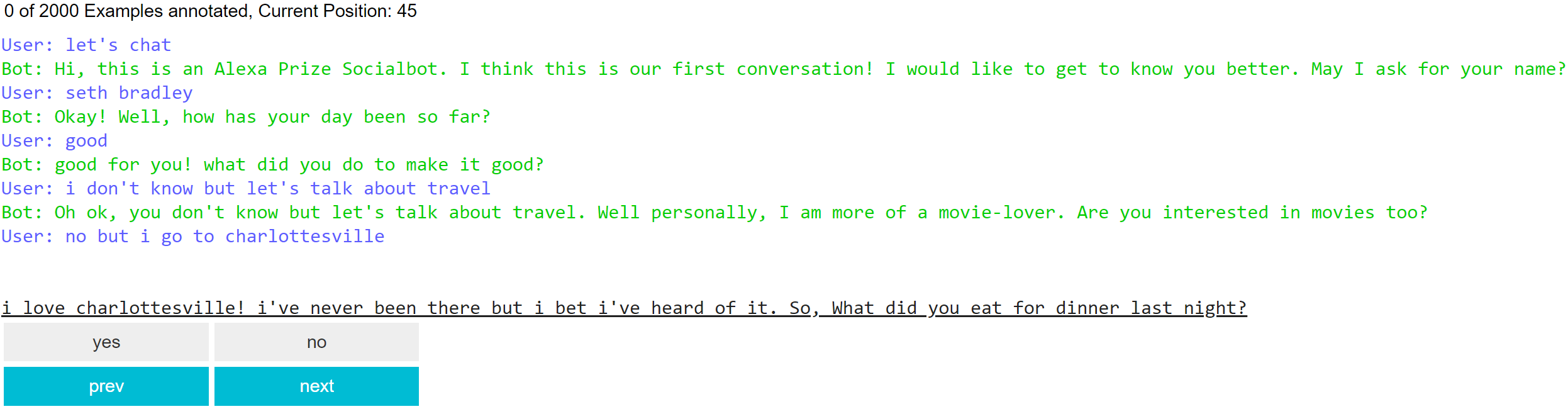}
    \caption{The annotation interface for our custom dataset. We keep it as simple as possible and use key bindings to make the annotation process fast.}
    \label{fig:annotation}
\end{figure*}

\textbf{Custom Dataset }
We create a dataset with selected conversations between users and our bot. 
For each bot turn, we keep all filtered candidate responses from our bot for annotation. The dataset consists of 965 conversations which include 8,357 bot turns in total, and these turns contain 60,281 candidate responses in total. 
The annotators are asked to label each candidate response as ``good" or ``bad" given the dialogue history, as shown in \autoref{fig:annotation}. 
They can also choose to skip labeling uncertain entries, and the skipped entries are then removed from the dataset. The total number of skipped entries after annotation is 942.
More detailed information about our annotated dataset is shown in \autoref{tab:data_statistics}.
The annotation interface is based on PigeonXT\footnote{\url{https://github.com/dennisbakhuis/pigeonXT}}, a simple open-source widget for data annotations on Jupyter notebooks. 

Note that in our annotations, factual correctness is not a factor that affects whether a response is good or bad.
Since the poly-encoder in its current form does not have access to a comprehensive information source for determining factual correctness, we do not burden the poly-encoder with such a role. 
Therefore, we annotate with the assumption that all given responses are factually correct and only determine whether the response is a suitable one for a friendly and unbiased socialbot to say to the given dialogue context. 
We impose the duty of factual correctness to the response generators themselves or additional guardrails. 


\begin{table}[h!]
\centering
\begin{tabular}{lrrr}
\hline
                                       & Total & Training & Validation  \\ \hline
No. of conversations                & 958   & 863      & 95 (43)     \\ 
No. of turns                        & 8,288  & 7,506     & 785 (488)  \\ 
No. of candidate responses          & 59,339 & 53,648    & 5,691 (5,394) \\ 
No. of candidates labeled as "good" & 30,888 & 27,904    & 2,984 (2,725)  \\ 
No. of candidates labeled as "bad"  & 28,451 & 25,744    & 2,707 (2,669) \\ \hline
\end{tabular}
\caption{The numbers of conversations, turns, candidate responses, and candidate responses with each label in our annotated dataset. The numbers after removing the turns with single candidates are shown in parentheses.}
\label{tab:data_statistics}
\end{table}

Besides the differences in domain and persona, our custom dataset has the following differences from ConvAI2: 
1) In the ConvAI2 dataset, each conversation history has only one ``good'' response, but the conversation histories in our custom dataset could have multiple or zero ``good'' responses. 
2) The ConvAI2 dataset does not contain dedicated distractor responses, but in our custom dataset, we have dedicated distractors which are the candidate responses labeled as ``bad'' for each conversation history.

\textbf{Fine-tuning: } We first introduce two sources of candidates during fine-tuning, which are ``batch'' and ``inline'' (these two terms are taken from the ParlAI library) \citep{miller-etal-2017-parlai}. 
``Batch'' means using the correct responses of other training examples in the same batch as distractors; 
``inline'' means that each training example comes with its own distractors. 
We format our custom dataset to have one training example for each response labeled as ``good'' in each turn and keep the responses labeled as ``bad'' as dedicated distractors when fine-tuning with inline candidates. 
To take advantage of the ConvAI2 dataset, we combine it with our custom dataset. 
Since we do not use persona information in our bot, we remove the personas in the ConvAI2 dataset to make it more consistent with our setting. 

We randomly pick 10\% of the conversations in our dataset for validation and fine-tune with the rest. 
Some turns in our dataset have only one candidate response.
We drop these turns from the validation set since selecting from a single response is meaningless, and keeping these turns makes the performance measurement inaccurate.
\autoref{tab:data_statistics} shows more details of the training and validation sets.
For the ConvAI2 dataset, we use the training-validation split from the ParlAI library.

\textbf{Fine-tuning with batch candidates:}
We first fine-tune the pre-trained poly-encoder on our custom dataset and the ConvAI2 dataset with batch candidates. 
During training, we use batches of size 20 and combine the two datasets by making each batch include 3 examples from our dataset and 17 examples from the ConvAI2 dataset. 
After our dataset runs out of examples, the remaining batches contain only examples from ConvAI2.
The distractors of each training example are correct responses of other examples in the same batch, so we also ensure that the 3 examples from our dataset in each batch are from different turns. 
We do not do this for ConvAI2 examples since each turn in the ConvAI2 dataset has only one corresponding training example.

\textbf{Fine-tuning with inline candidates:}
We also fine-tune a poly-encoder model on both dataset with inline candidates. We use batches of size 10, and each training example comes with 9 dedicated distractors. 
For the training examples with more than or less than 9 dedicated distractors, we randomly select 9 instances from the dedicated distractors or randomly select distractors from the response pool of the ConvAI2 training set to make up the difference, respectively. 
Compared with the ``batch'' setting, the ``inline'' setting is less memory efficient because of the dedicated distractors, but the model is aware of the real distractors that are generated with same conversation histories.

%% file: sections/analysis_findings.tex
\section{Analysis and Findings}
\label{sec:analysis}

\subsection{Ratings by Response Generator}

\begin{figure*}[ht]
    \centering
    \includegraphics[width=\textwidth]{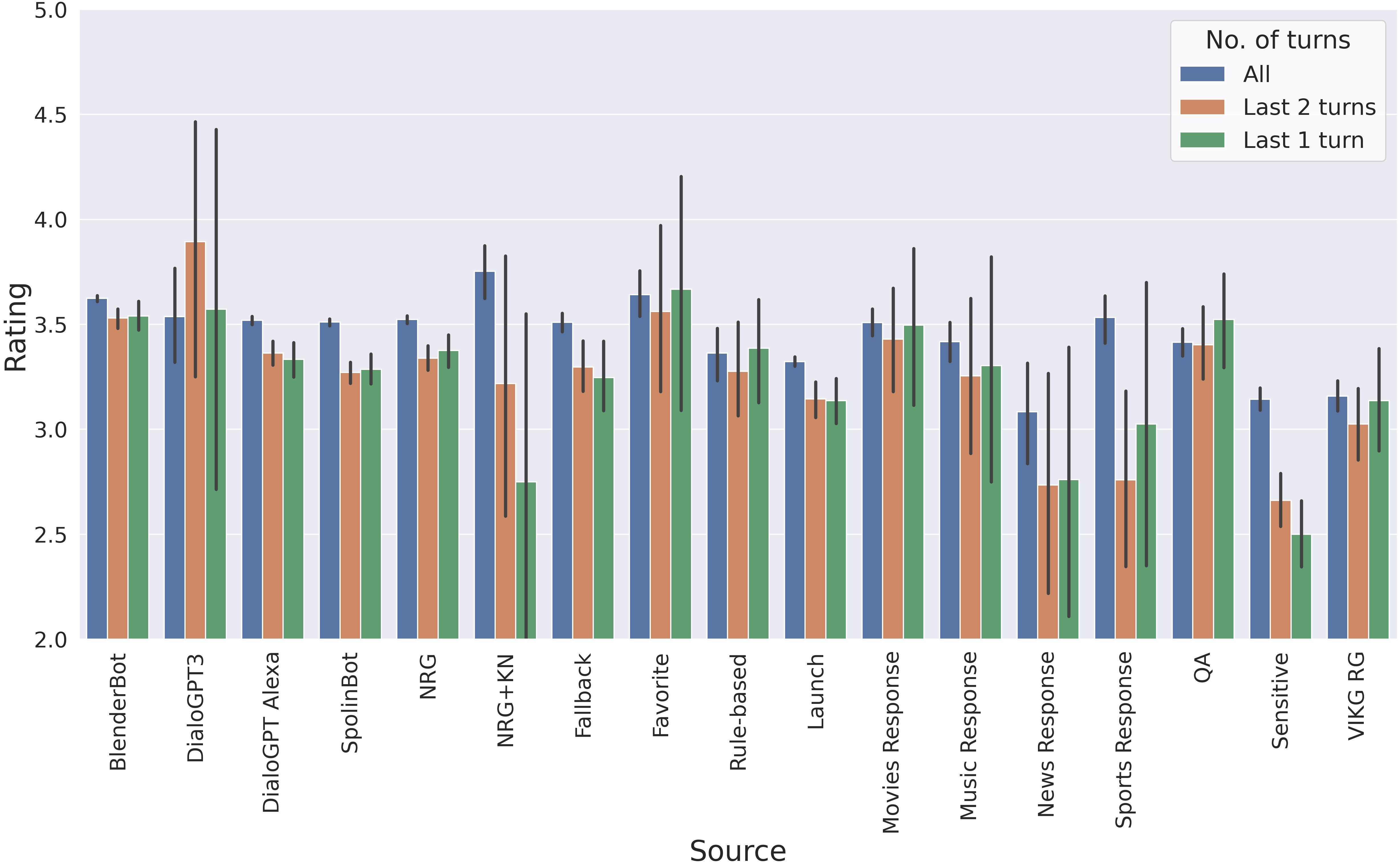}
    \caption{Average ratings of response generators throughout the semifinals when assigning the conversation ratings to response generators for all turns, last 2 turns, and only the last turn before the stop intent.}
    \label{fig:ratings_by_rg_1}
\end{figure*}

\begin{figure*}[ht]
    \centering
    \includegraphics[width=\textwidth]{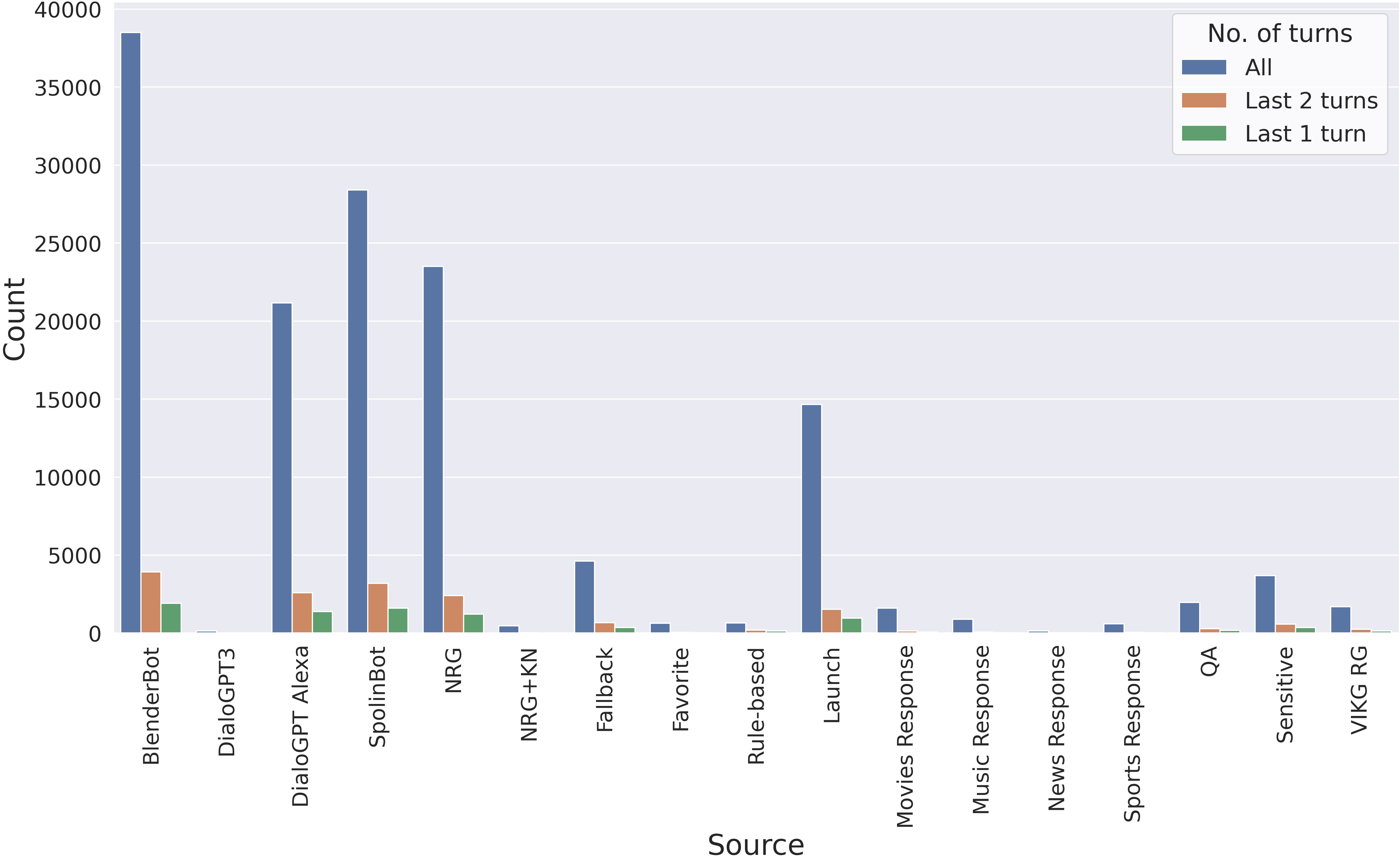}
    \caption{Number of turns contributed to the ratings of each response generator throughout the semifinals when assigning the conversation ratings to response generators for all turns, last 2 turns, and only the last turn before the stop intent.}
    \label{fig:ratings_by_rg_2}
\end{figure*}

For each conversation, we identify the response generator whose response was selected for each turn and then we identify the rating that this conversation has received. We determine this correlation for all the conversations during the semifinals duration in Figure \ref{fig:ratings_by_rg_1}. At the same time, we try to understand insights from this data by plotting the number of turns where the corresponding response generator was selected from Figure \ref{fig:ratings_by_rg_2}. In both these figures, we construct bar charts for all turns, last 2 turns and last 1 turn before the stop intent of each conversation. We plot separate data points for last $n$ turns based on the intuition that the last few turns contribute most to the overall rating. 

As shown in \autoref{fig:ratings_by_rg_2}, most of the responses came from Blenderbot, followed by SpolinBot and then NRG. The conversations these contributed to received fairly high ratings. At the same time, though DialoGPT3 and Favorite modules' responses were selected very few times, they contributed to high-rated conversations. Ratings of conversations where domain-based response generators (Movies, Music, Sports) were used are also fairly high. The News domain-based response generator had poor performance and so it was dropped during the semifinals.

Neural-based response generators have contributed to the last 2 turns in more conversations than  domain-specific ones have. Generally, conversations involving domain-based topics completed and then transitioned to general utterances before the conversation ended.



\subsection{Latency reduction and Latency vs Ratings Correlation}



\begin{figure*}[ht]
    \centering
    \includegraphics[width=\textwidth]{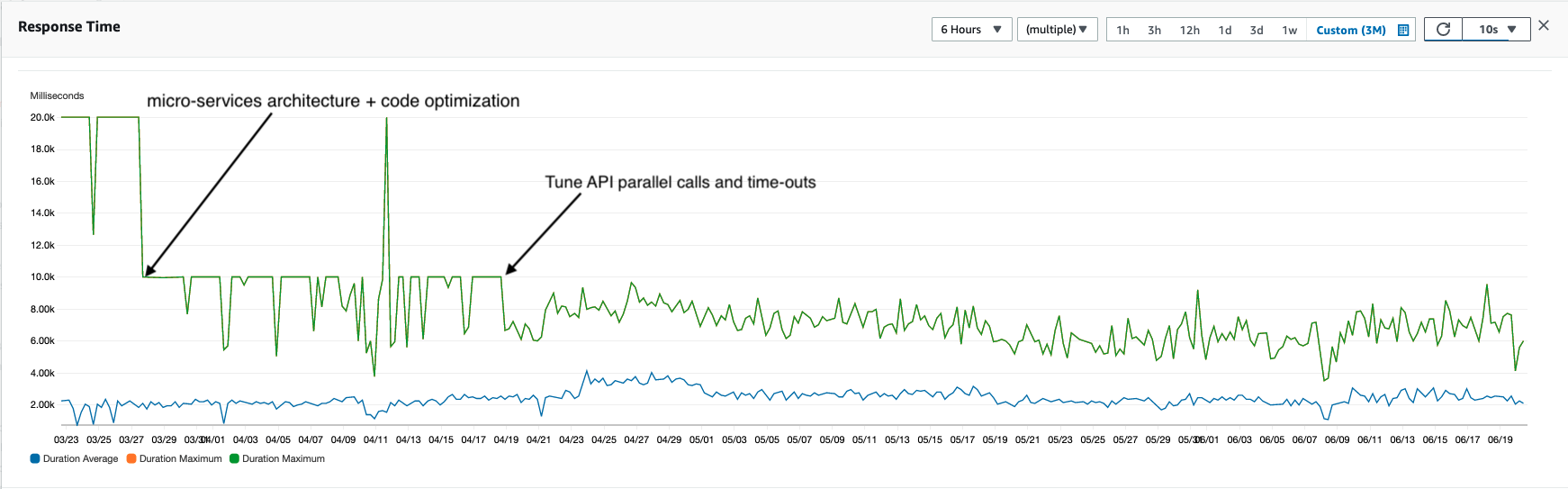}
    \caption{Overall response time for the last 3 months}
    \label{fig:responsetime_lambda}
\end{figure*}

As we add response generators and evaluators to the pipeline, managing processing time becomes of primal importance. We experienced many low ratings in the third week of March, which we mainly attribute to longer conversation turns. We occasionally exhausted the time needed to process inputs, and thus maxed out the 20 second timeout set in AWS Lambda, as shown in  Figure \ref{fig:responsetime_lambda}.

We have to simultaneously ensure that system components produced outcomes of sufficient quality while reducing the impact of latency on the conversation experience. By adopting the micro-services architecture that is common in today’s backend systems, we can orchestrate all the micro-services from a single service and concurrently call remote modules. Starting with code optimization for feature extractors and response generators, by the end of March. we reduced all response times to less than 10 seconds, as shown in  Figure \ref{fig:responsetime_lambda}.

\begin{figure*}[ht]
    \centering
    \includegraphics[width=\textwidth]{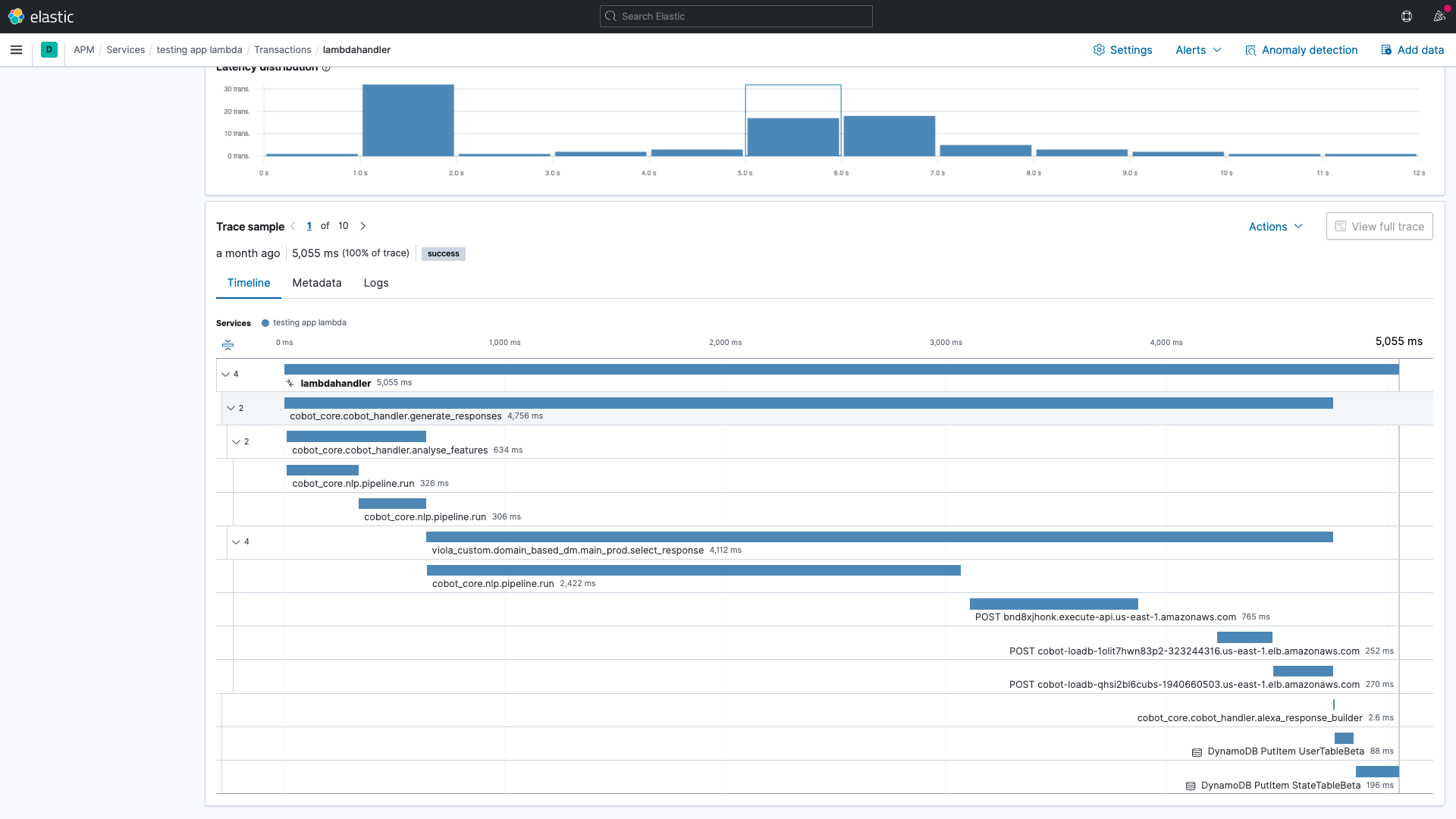}
    \caption{A breakdown of a single turn from Viola shown in the Elastic APM transaction distribution.}
    \label{fig:elastic_apm}
\end{figure*}

We employ Application Process Monitoring (APM) from Elastic to visualize transactions together with a distributed trace to get a comprehensive view of how services interact. 
We can see the transaction breakdown for a single request in Figure \ref{fig:elastic_apm}. 
It is helpful to examine service calls across the modules, discover where latency issues are arising in the program, and pinpoint the components that need optimizing. 
We tune attributes like API read-timeouts for remote modules such as NRG based on the performance impact on the request.

To further boost performance for modules with great variability in latency, we double the number of parallel executions and wait for half of the invocations to be completed before proceeding. This reduces the chances of being bottle-necked by a slower invocation. It helps us reduce the average response time for the Topical NRG and PD-NRG modules by almost 500ms.


\begin{figure*}[ht]
    \centering
    \minipage{0.32\textwidth}
    \includegraphics[width=\textwidth]{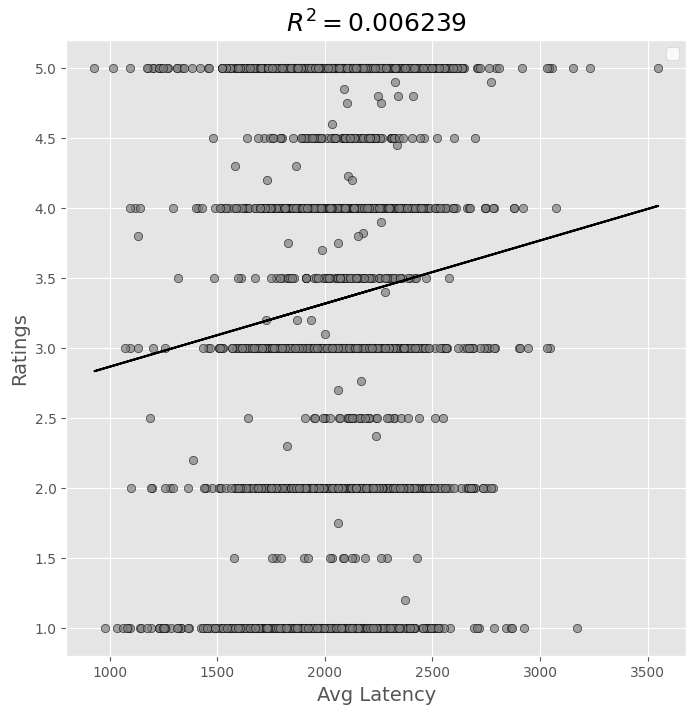}
    \caption{Avg Latency vs. Rating for all turns}
    \label{fig:r2_latency_rating}\endminipage\hfill
    \minipage{0.32\textwidth}
    \includegraphics[width=\textwidth]{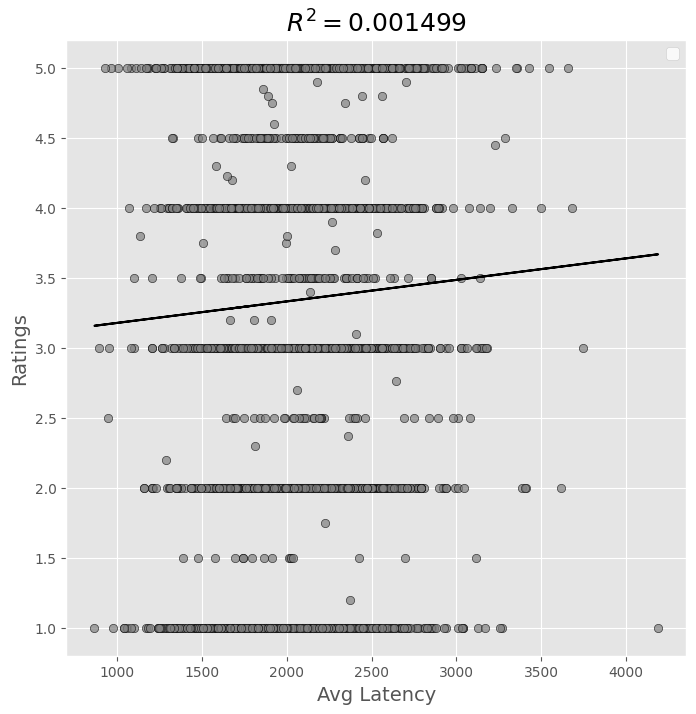}
    \caption{Avg Latency vs. Rating for last turn}
    \label{fig:r2_latency_rating_1n}\endminipage\hfill
    \minipage{0.32\textwidth}
    \includegraphics[width=\textwidth]{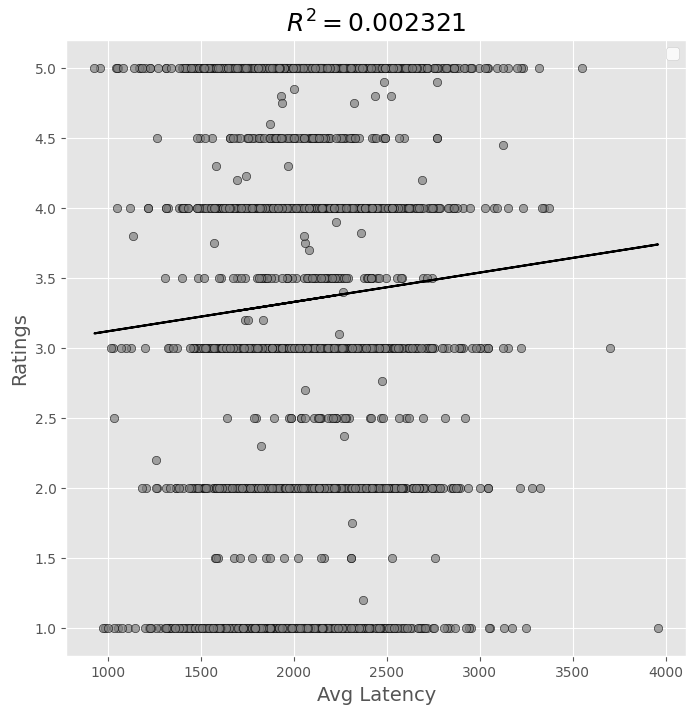}
    \caption{Avg Latency vs. Rating for last tow turns}
    \label{fig:r2_latency_rating_2n}\endminipage\hfill
\end{figure*}

While examining the relationship between ratings and latency, we observe no noticeable relationship between the two variables when calculating the R-Squared correlation value, as seen in Figure \ref{fig:r2_latency_rating}. A similar trend can be observed in the R-Squared correlation value for last one and two conversation turns in Figure \ref{fig:r2_latency_rating_1n} and Figure \ref{fig:r2_latency_rating_2n}, respectively. Hence, we see that latency does not significantly affect user ratings if none of the responses time out at 10 seconds and if the conversation's average latency is less than 3.5 seconds.

\subsection{Experiments with the response selector}

\begin{table}[h!]
\centering
\begin{tabular}{lrr}
 \hline 
Model                                 & ConvAI2        & Our Dataset    \\ \hline
Baseline (random guess)          & 5.00           & 54.86          \\ 
Conversation evaluator           & 8.17           & 57.30          \\ 
Default poly-encoder             & 66.50          & 58.19          \\ 
Fine-tuned poly-encoder (batch)  & \textbf{70.25} & 59.07          \\ 
Fine-tuned poly-encoder (inline) & 68.18          & \textbf{69.47} \\ \hline
\end{tabular}
\caption{Accuracy of each model on the validation set of the ConvAI2 dataset and our custom annotations.}
\label{tab:selector_scores}
\end{table}

\begin{figure}
     \centering
     \begin{subfigure}[b]{\textwidth}
         \centering
         \captionsetup{skip=0pt}
         \includegraphics[width=\textwidth]{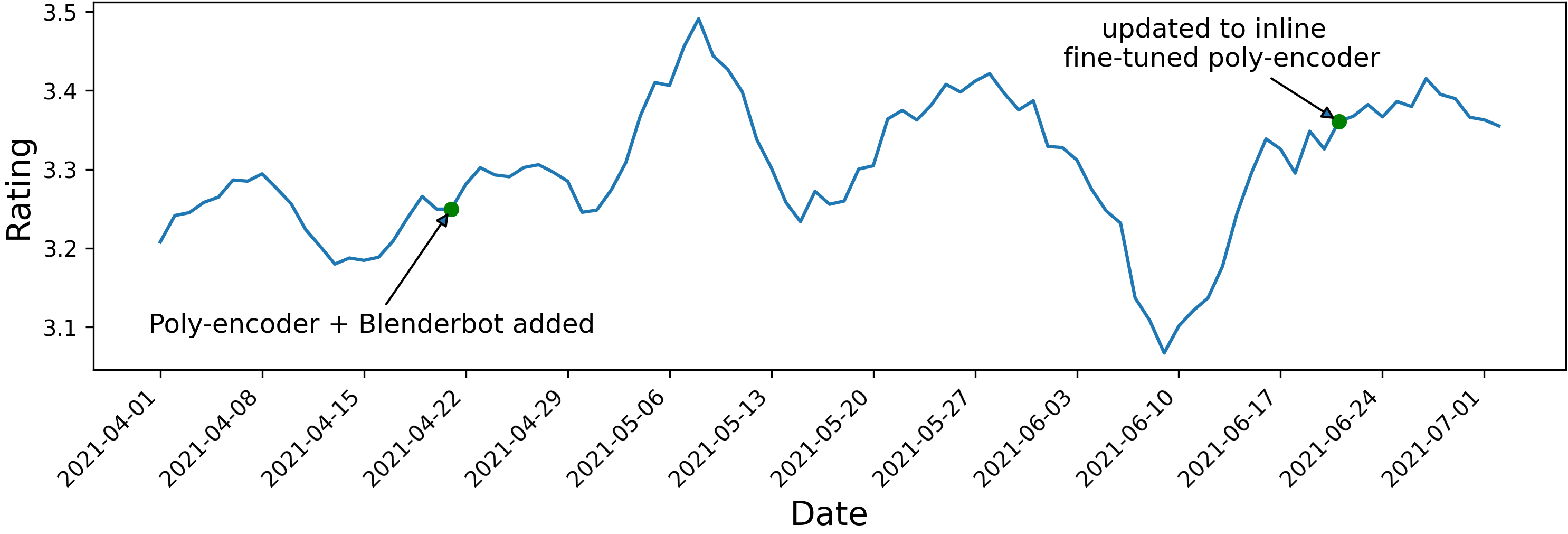}
         \caption{Main traffic}
         \label{fig:conv_eval2a}
     \end{subfigure}
     \par\medskip
     \par\medskip
     \begin{subfigure}[b]{\textwidth}
         \centering
         \captionsetup{skip=0pt}
         \includegraphics[width=\textwidth]{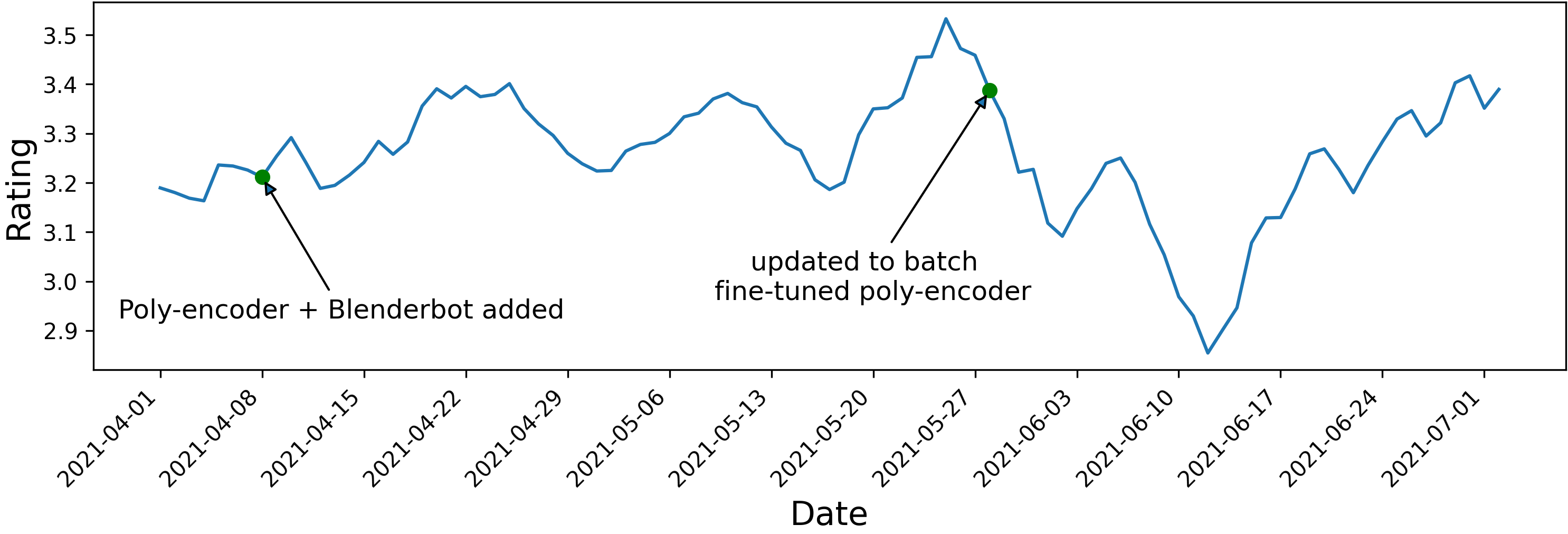}
         \caption{Experimental traffic}
         \label{fig:conv_eval2b}
     \end{subfigure}
        \caption{Rolling 7-day average rating curve for main and experimental traffic. The most notable changes to the selectors are indicated by arrows and green dots.}
        \label{fig:conv_eval2}
\end{figure}


\subsubsection{Conversation Evaluator vs. Poly-encoder}

One of the major experiments that we conducted for Viola is the shift from the default conversation evaluator to the polyencoder as a selector. 
The conversation evaluator assesses a response across multiple dimensions, but it is challenging to pick a single dimension or combination of dimensions that aligned most with our judgement. 
Also, it is a default API provided as part of Cobot that can not be retrained with our own data.

We compare the performance of the conversation evaluator' \texttt{is\_erroneous} metric and the default poly-encoder on the validation set of the ConvAI2 dataset and our custom dataset to make a direct comparison between them. As shown in \autoref{tab:selector_scores}, the conversation evaluator performs a little better than the baseline on both validation sets, and the default poly-encoder has a slightly better accuracy than the conversation evaluator on our validation set but is much better on the ConvAI2 validation set. One obvious reason is that the default poly-encoder is fine-tuned on the ConvAI2 dataset. Also, we use the default setting of the conversation evaluator, which only keeps two previous turns as the context. The poly-encoder model keeps all turns,\footnote{We remove the oldest history utterances from the conversation history if the total length exceeds the maximum input length of the transformers, which in our case is 1024 tokens.} so it can capture more context information than the conversation evaluator.

One may notice that the baseline scores are very different on the two datasets. The reason is that the score on the ConvAI2 dataset is hits@1/20, meaning exactly 1 out of the 20 candidates is correct for each example, while the score on our dataset is hits@1/(number of candidates), where the number of candidates for each example is usually less than 20, and in most cases, more than one candidate can be a good response. 
More specifically, as shown in \autoref{tab:data_statistics}, 50.52\% (2,725 out of 5,394) candidate responses in our validation set are ``good'' responses, so the baseline can achieve over 50\% accuracy on our validation set with random guessing.
It is also worth noticing that the accuracy of the default poly-encoder on the ConvAI2 validation set is much lower than the 89+\% accuracy mentioned earlier. That is because the default poly-encoder model is fine-tuned on data with personas, but we remove personas from the ConvAI2 dataset for our purpose.

Based on these results, we test the off-the-shelf poly-encoder weights that are trained with ConvAI data in the experimental traffic. 
As shown in \autoref{fig:conv_eval2}, we saw an increase in ratings for the experimental traffic and decided to use it for the main traffic as well, which also led to a rise in the main traffic ratings.
The 7-day rolling average rating increases from 3.250 to 3.296 on the main traffic and increases from 3.211 to 3.241 on the experimental traffic in 7 days after the transitions.
The results are not well controlled as changes in ratings during the time frame shown in the figure were also affected by engineering faults that led to downtime of certain response generators. 
However, the overall positive trend among the conversations that did not have engineering faults encouraged us to proceed to fine-tuning the poly-encoder with our own data such that it learns to make choices that align with our choices.

\subsubsection{Fine-tuned Poly-encoder}

The accuracy of each fine-tuned poly-encoder model is shown in \autoref{tab:selector_scores}. Both fine-tuned poly-encoder models outperform the default poly-encoder on both validation sets. 
The poly-encoder fine-tuned with batch candidates only improves a little over the default poly-encoder, but the one fine-tuned with inline candidates has an improvement of over 10\%.
A possible reason for this is that the inline fine-tuned poly-encoder takes advantage of dedicated distractors, which are responses generated with the same context as the correct response. This model is aware of the real bad responses of the given context and can assign lower scores to these responses or others like them during inference. 
This can potentially increase the prediction margin between the ``good'' and the ``bad'' responses. 
The batch fine-tuned poly-encoder is mainly optimized for the ``good'' responses only. 
We also believe that the poly-encoder fine-tuned with inline candidates would perform better on the real responses generated by our bot because the fine-tuning process is closer to the setting of real responses.

We also test both versions of the fine-tuned poly-encoder in the experimental and main traffic. 
As shown in \autoref{fig:conv_eval2}, after updating to the batch fine-tuned poly-encoder, the rating on the experimental traffic drops rapidly. 
Since the rating has already started dropping a few days before the update, we cannot solely blame the batch fine-tuned poly-encoder for the drop. 
It can be that the experimental traffic's ratings are more volatile due to the smaller sample size. 
Nevertheless, the ratings continue to drop well after the transition, so we can infer that this transition to the batch fine-tuned poly-encoder was not helpful for the ratings. 
In contrast, from \autoref{fig:conv_eval2}, we can see that after updating to the inline fine-tuned poly-encoder, the rating on main traffic increases. 
Specifically, the 7-day rolling average rating increases from 3.361 to 3.395 in 7 days. 
The changes in ratings are consistent with the test results on our validation set where the batch fine-tuned poly-encoder has very little improvement over the default poly-encoder, while the inline fine-tuned poly-encoder outperforms the default poly-encoder significantly.




\subsection{Speech sentiment score vs Ratings}

\begin{figure*}[ht]
    \centering
    \minipage{0.31\textwidth}
    \includegraphics[width=\textwidth]{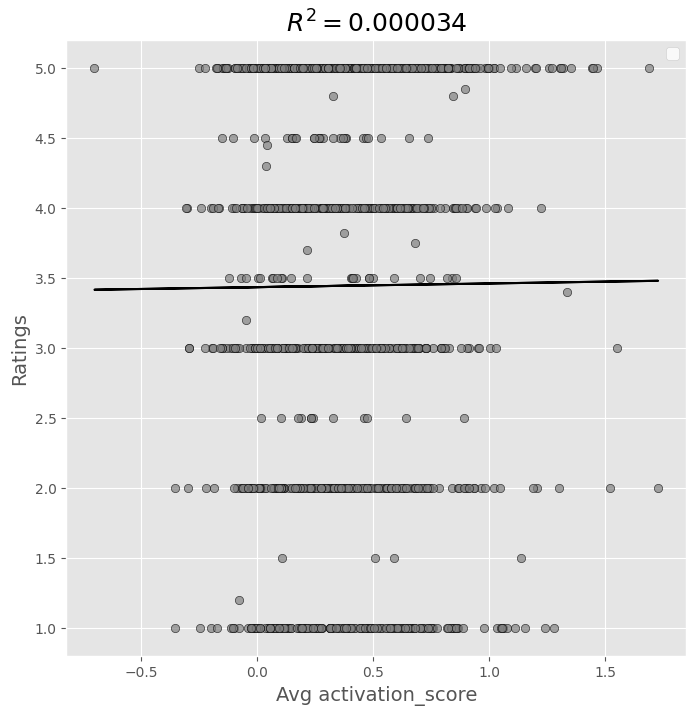}
    \caption{Avg. Activation Score vs. Rating}
    \label{fig:activation_score_avg}\endminipage\hfill
    \minipage{0.31\textwidth}
    \includegraphics[width=\textwidth]{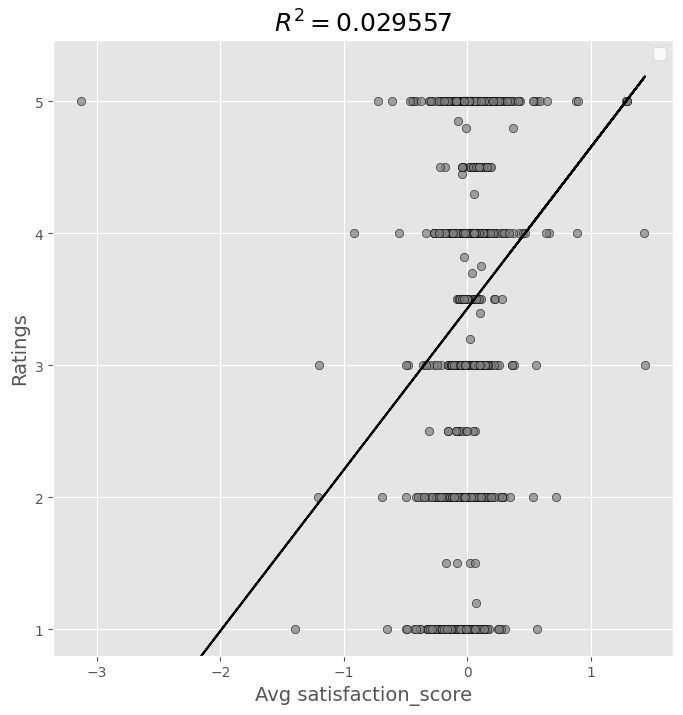}
    \caption{Avg. Satisfaction Score vs. Rating}
    \label{fig:satisfaction_score_avg}\endminipage\hfill
    \minipage{0.31\textwidth}
    \includegraphics[width=\textwidth]{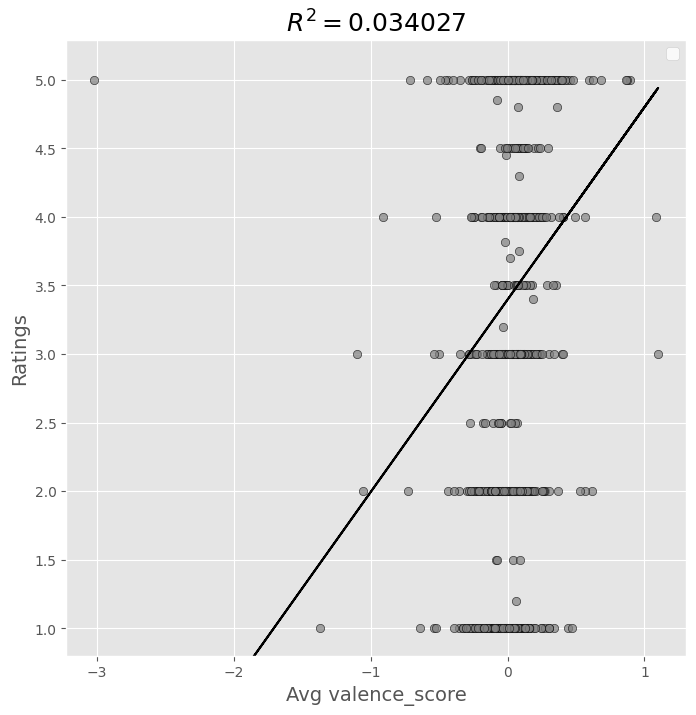}
    \caption{Avg. Valence Score vs. Rating}
    \label{fig:valence_score_avg}\endminipage\hfill
\end{figure*}

With the newest update to Cobot, we are able to retrieve audio-level sentiment scores for each turn. The sentiment score is given in three dimensions: satisfaction, activation, and valence.
Satisfaction measures the degree of positivity/negativity toward Alexa. Low satisfaction (-3) indicates extreme frustration, whereas high satisfaction (+3) indicates delight.
Activation measures the intensity of an emotion. High activation (+3) can arise from someone shouting from excitement or anger, whereas low activation (-1) can arise if someone is subdued or tired sounding.
Lastly, valence measures the degree of positivity/negativity of an emotion in general. Low valence (-3) can indicate extreme frustration or sadness, whereas high valence (+3) indicates happiness. 
This was an exciting new feature to have for generating responses that incorporated information that was embedded directly in speech. 
We wanted to determine whether these scores were correlated to conversation ratings to assess the viability of using these features as turn level indicators for user satisfaction.

However, from Figures \ref{fig:activation_score_avg},\ref{fig:satisfaction_score_avg}, and \ref{fig:valence_score_avg}, we can see no correlation between the Sentiment Activation, Sentiment Satisfaction or Valence scores and user ratings. These figures compare user-provided scores with average predicted sentiment scores across an entire conversation, and excluded outliers. 
We also tried using the maximum or minimum scores (under the assumption that a very good or very bad turn might be remembered) as well as limiting the range of turns in the conversation over which the average, max, or min was taken to be the final turn or up to the final three turns. This was done under the motivation that users are most affected by their experience immediately before they provide a rating. 
However, we saw nearly the same uncorrelated results in all these variants.
 This is a preliminary analysis, since the speech sentiment scores only became available during the last 12 days of the semifinals. As our system was mostly stable\footnote{Other than a time frame in which an updated ASR service classified the word `stop' as `stapp' and increased user's frustration.} over this period, it is possible there was not enough noticeable difference in the experiences to be detected by the sentiment models.


%% file: sections/discussion_future_work.tex
\section{Discussion and Future Work}

\subsection{Data distribution shift between training time and deployment}

While generative models for open-domain dialogue have come a long way, we observe many conversations for which they come up short, mainly due to the data distribution shift between training time and deployment. 
Although neural models are touted for their generalizability, it is usually within relatively narrow domains that they perform well without hand-crafted rules.
Even datasets for open-domain dialogue that are meant to be topic agnostic are collected with a specific skillset in mind, such as being empathetic or grounding the conversation to a given persona. This narrows down the functionality of the models trained on them.

The most critical limitation of the datasets used for Viola’s generative RGs is that they do not capture many of the circumstances that occur for a speech-based socialbot facing public users. Training data mostly contains conversations that are typed and are between well-behaved human beings that more or less have equal status and capabilities, but speech-based socialbots must understand noisy ASR output, often encounter capricious and malicious users, and are not yet treated as nor assumed to be equals. 

\textbf{Typed text vs. ASR output: } The first major shift from training to deployment is that models need to handle noisy ASR output instead of typed text. 
A user typing can take a long time to formulate a response and fix any errors. Hence, typed conversations contain very few nonsensical utterances. Typed conversations also have the advantage of including punctuation, which clarifies any ambiguities. Spoken conversations often contain nonsensical input and never contain punctuation.  
As human beings, the most natural response to nonsensical utterances would be to question their validity and ask for clarification, but neural models are ill-prepared to handle these cases because they have always seen utterances that made sense and will always attempt to try to respond to them assuming they make sense. 

Another issue with spoken utterances is that they tend to be shorter than typed text. 
In the Alexa Prize Challenge, Alexa will allow a limited time span for the user to reply and will interpret a long break after the user starts talking as a signal that the user has finished talking. When we were internally testing Viola via an Alexa device, we often found ourselves rushed to have a well-formulated response and say it in one go or have our utterance truncated to when we took a break to think. 
Users are quick to realize the limits of this modality and thus they resort to short utterances in order to increase the likelihood that they will be understood by Alexa. 
Even as human beings, terse replies often make it difficult to continue a conversation, and given that there are not many conversation instances in the training data with this imbalance, it is not really a surprise that trained models tend to struggle with short utterances. 

\textbf{Human-human vs. human-bot: } In the training data, conversation participants are incentivized to behave well: crowdsourced workers follow rules for high quality data collection to ensure they are paid and Reddit users make their point clear in order to contribute to a conversation that they are genuinely interested in. 

On the other hand, as in the case of Microsoft Tay, we observe many real users that abuse socialbots and converse with them with the sole purpose of making them say problematic things. 
Although conversation threads from Reddit contain utterances coming from adversarial users and other users’ responses to them, the responses may also be offensive and not suitable for a socialbot to learn from. 
Determining what is the best way to converse with adversarial users is a subjective decision with no single right answer, but nonetheless it is something socialbots need to learn how to do that is not present in the current training data.  

Training socialbots with human-human data is also problematic simply because bots are not humans. If we naively train bots with human-human data, it will learn to say things that only apply to humans. Since users generally know that bots are not generally advanced enough to be able to do human activities such as eating or going shopping, bot utterances that presume otherwise are jarring to users and yield negative experiences. 

\subsection{Bridging the shift: proposed data annotation pipeline}

In light of the renewed attention on the importance of data quality over developing larger and more complex models \citep{andrewng_dataisfoodforai, bender2021dangers}, we believe the most effective way to improve socialbots is by iteratively expanding a dataset that bridges the aforementioned shift through a scalable quality-oriented data annotation pipeline. 

First, the pipeline should detect conversations that have utterances that are either repetitive and terse, not understandable to humans even with the dialogue context, or express dissatisfaction or confusion. 
We can use simple rules or train custom bootstrapped classifiers by annotating conversations with low ratings to detect various classes of user utterances that provide useful signal. If correlated with user experience, the newly available turn-level sentiment scores can also be useful to model fine-grained satisfaction feedback and lead to better-informed model modifications. 
Annotations can be done simultaneously with the poly-encoder annotations, and therefore they can be done quickly with our annotation UI. 
With the detected results, the pipeline should facilitate the provision of alternative suitable socialbot responses for these instances. This can be done manually or with automatic methods such as our DialoGPT + GPT-3 approach. 
Iterating through this pipeline, we will be able to quickly deal with frequently occurring cases and find edge cases, creating a dataset that lays out the foundation for a robust socialbot.

\subsection{Advanced knowledge retrieval and factual correctness}

Making sure that the socialbot provides engaging, coherent, and factually correct responses grounded to relevant and updated information is a challenging task not addressed with the data annotation pipeline. By taking a Wizard of Oz approach and taking on the role of a socialbot, we realized the limits of Viola's current knowledge retrieval approach as we rely on the entire dialogue history rather than only the detected entities or the current utterance when we retrieve the most relevant pieces of knowledge for grounding our responses. 
Based on this observation, we need to update our knowledge retrieval module to become more versatile, understanding from the conversation what kind of information might be useful for the conversation and be more interesting without an explicit query or trigger. We also need to make sure that the information is truthful in order to guide RGs to generate factual responses.

\subsection{Long-term memory and personalization}

Since having a good first conversation with a user is already a challenging task, we did not focus on approaches that scale to multiple conversations with the same user and maintain a sophisticated memory architecture. 
With the lack of conversation data between two parties that span over several separate occasions, it is challenging to test models that are able to leverage long-term memory to ground responses, let alone train them. 
The first step in this direction is to construct a small dataset that tests this long-term memory capacity. 
Once the data is ready, it would be interesting to experiment with models that can attend over very long sequences and models that maintain separate weights for memory and language generation capacity \citep{beltagy2020longformer, yogatama2021adaptive}. 




%% file: sections/appendix.tex
\include{assets/gpt3_conversations}

%% file: assets/gpt3_conversations.tex
\begin{table}
    \centering
\rowcolors{1}{white}{gray!20}
\renewcommand{\arraystretch}{1.1}
\resizebox{13.2cm}{!}{ 
  \begin{tabular}{ | m{3.2cm} | p{10cm} | }
    \hline
    \textbf{Source} & \textbf{Utterance} \\ \hline
    USER & let’s chat \\ \hline
    VIOLA & Hi, this is an Alexa Prize Socialbot. I think this is our first conversation! I would like to get to know you better. May I ask for your name? \\ \hline
    
    USER & janice \\ \hline
    VIOLA & Nice to meet you janice! How has your day been so far? \\ \hline
    
    USER & not so great \\ \hline
    VIOLA (\textcolor{blue}{GPT-3}) & How dare you say that on our banner! (\textcolor{blue}{Oh no! I am sorry to hear that. Do you want to talk about it?})\\
    \hline
  \end{tabular}
  }
  
\resizebox{13.2cm}{!}{ 
  \begin{tabular}{ | m{3.2cm} | p{10cm} | }
    \hline
    \textbf{Source} & \textbf{Utterance} \\ \hline
    ... & ... \\ \hline
    VIOLA & I am glad to hear that. What was your day like? \\ \hline
    
    USER & i want to learn about jesus and religion \\ \hline

    VIOLA (\textcolor{blue}{GPT-3}) & Ok. Do you like being alone? (\textcolor{blue}{Interesting. I've learned the Bible and would like to discuss it with you.})\\
    \hline
  \end{tabular}
  }
    \caption{Examples where GPT-3's alternative responses provided a clear improvement over the original.}
    \label{tab:gpt3-good}
\end{table}

\begin{table}
    \centering
\rowcolors{1}{white}{gray!20}
\renewcommand{\arraystretch}{1.1}
\resizebox{13cm}{!}{ 
  \begin{tabular}{ | m{3.2cm} | p{10cm} | }
    \hline
    \textbf{Source} & \textbf{Utterance} \\ \hline
    ... & ... \\ \hline
    VIOLA & I saw online that chicken is a subspecies of the red junglefowl.. Wanna know something else? \\ \hline
    
    USER & anything other than chicken \\ \hline
    
    VIOLA (\textcolor{red}{GPT-3}) & It's about the guy who got eaten by an ant.  (\textcolor{red}{Do you like to talk about chicken?}) \\
    \hline
  \end{tabular}

  }
  
\resizebox{13cm}{!}{ 
  \begin{tabular}{ | m{3.2cm} | p{10cm} | }
    \hline
    \textbf{Source} & \textbf{Utterance} \\ \hline
    ... & ... \\ \hline
    VIOLA & What do you mean? You don't like the way she sounds when she sings? \\ \hline
    
    USER & no what \\ \hline
    
    VIOLA (\textcolor{red}{GPT-3}) & That's fine, I have lots of other interests too. so um, I sort of have a confession... I like to watch funny animal videos when I’m supposed to be learning new things. Maybe you can relate! Do you like animals too?  (\textcolor{red}{I'm confused.}) \\
    \hline
  \end{tabular}

  }
    \caption{Examples where GPT-3's alternative responses were not preferable over the original response.}
    \label{tab:gpt3-bad}
\end{table}